\documentclass[runningheads]{llncs}

\usepackage[mobile]{eccv}

\usepackage{eccvabbrv}

\usepackage{graphicx}
\usepackage{booktabs}

\usepackage[accsupp]{axessibility}  % Improves PDF readability for those with disabilities.

\usepackage{xparse}
\usepackage{cite}
\usepackage[resetlabels,labeled]{multibib}
\newcites{S}{References for Supplement}
\newcites{M}{References}
\newcommand{\citeApp}[1]{\citeS{#1S}}
\usepackage{hyperref}

\usepackage{orcidlink}

\input{preamble}

\begin{document}

\title{Good Teachers Explain: Explanation-Enhanced Knowledge Distillation} 


\author{Amin Parchami-Araghi$^{*}$\orcidlink{0000-0003-0424-7812} \and
Moritz Böhle$^{*}$\orcidlink{0000-0002-5479-3769} \and
Sukrut Rao$^{*}$\orcidlink{0000-0001-8896-7619} \and
Bernt Schiele\orcidlink{0000-0001-9683-5237}}

\authorrunning{A.~Parchami-Araghi et al.}

\institute{Max Planck Institute for Informatics, Saarland Informatics Campus, Saarbrücken \\
\email{\{mparcham,mboehle,sukrut.rao,schiele\}@mpi-inf.mpg.de}
}
% \and
% Springer Heidelberg, Tiergartenstr.~17, 69121 Heidelberg, Germany

\maketitle
\begin{center}
\begin{adjustbox}{minipage=0.94\linewidth, scale=1}
    \centering
    \captionsetup{type=figure}
    \includegraphics[width=0.6\textwidth]{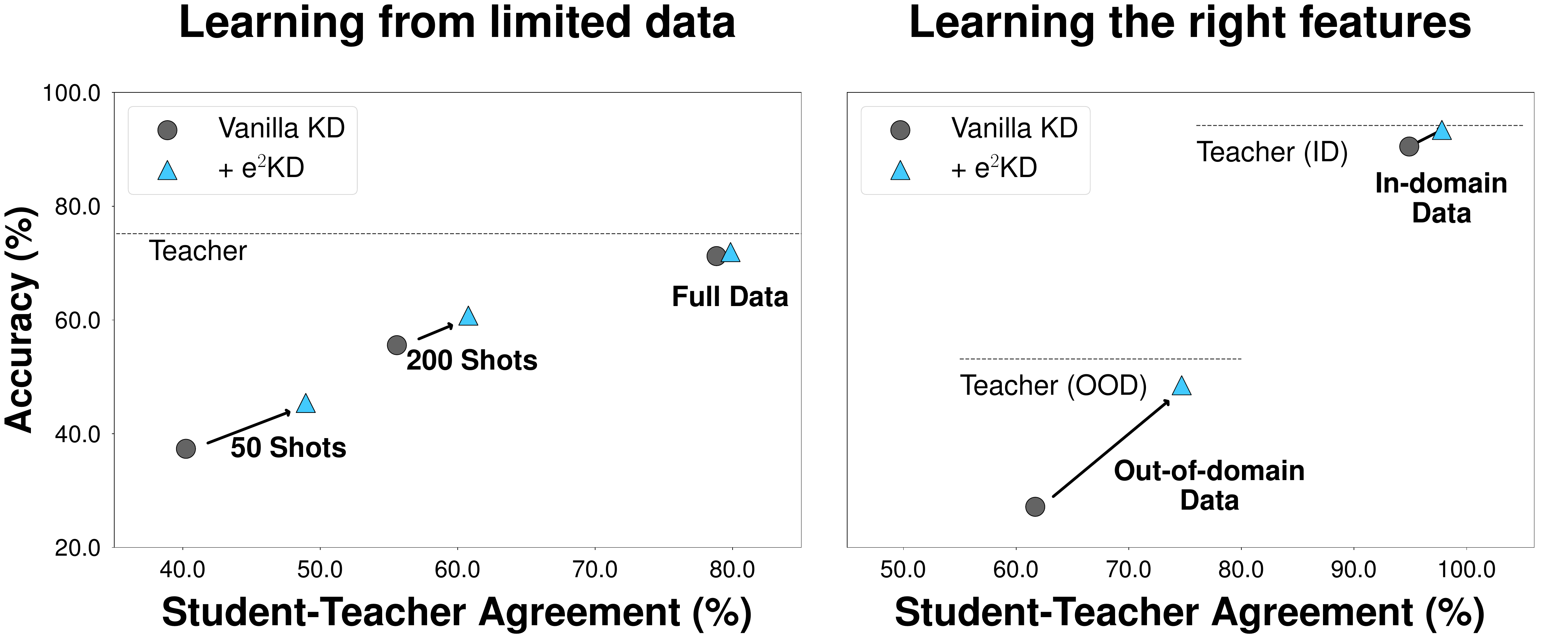}%
\includegraphics[width=0.4\textwidth]{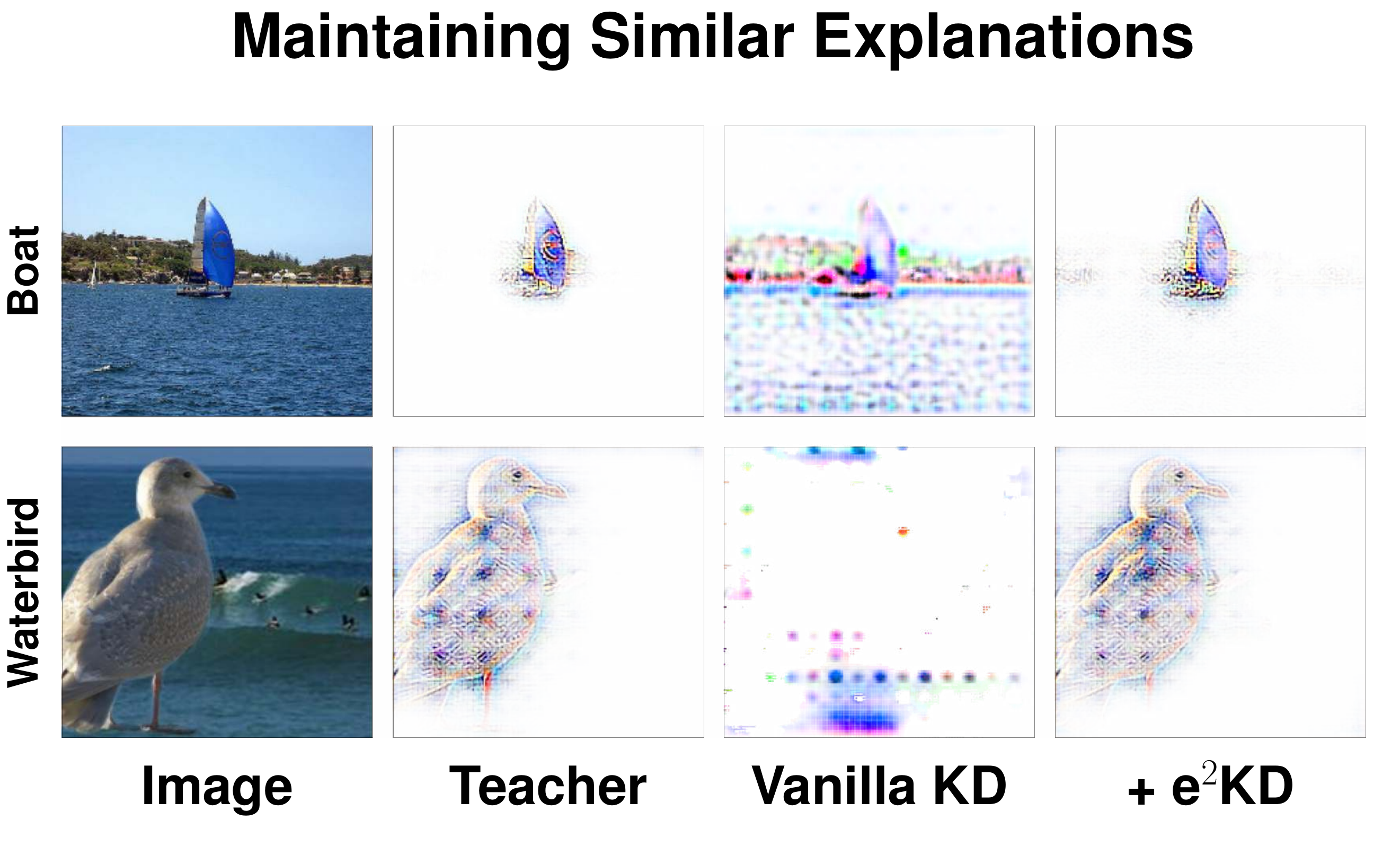}
\caption{\textbf{A good teacher explains.}
Using explanation-enhanced KD (\textbf{\eKD}) improves distillation faithfulness and student performance. \Eg, \eKD allows the student models to more faithfully approximate the teacher, especially when using fewer data, leading to large gains in accuracy and {teacher-student agreement}
 \textbf{(left)}. Further, by guiding the students to give similar explanations 
as the teacher, \eKD ensures that students learn to be `right for the right reasons', improving their accuracy under distribution shifts \textbf{(center)}. Lastly, \eKD students learn similar explanations as the teachers, thus exhibiting a similar degree of interpretability as the teacher \textbf{(right)}.
}
  \label{fig:final-teaser}
\end{adjustbox}
\vspace{-0.1cm}
\begin{abstract}\noindent
Knowledge Distillation (KD) has proven effective for compressing large teacher models into smaller student models. While it is well known that student models can achieve similar accuracies as the teachers, it has also been shown that they nonetheless  often do not learn the same function. 
It is, however, often highly desirable that the student’s and teacher’s functions share similar properties such as basing the prediction on the same \features, as this ensures that students learn the `right features' 
from the teachers. In this work, we explore whether this can be achieved by not only optimizing the classic KD loss but also the similarity of the explanations generated by the teacher and the student. 
{Despite the idea being simple and intuitive}, we find that our proposed `explanation-enhanced' KD (\eKD) (1) consistently provides large gains over logit-based KD in terms of accuracy and student-teacher agreement, (2) ensures that the student learns from the teacher to be right for the right reasons and to give similar explanations, and (3) is robust with respect to the model architectures, the amount of training data, and even works with `approximate', pre-computed explanations.
\end{abstract}
{}
\end{center}

\def\thefootnote{*}\footnotetext{Denotes equal contribution. Code: \href{https://github.com/m-parchami/GoodTeachersExplain}{github.com/m-parchami/GoodTeachersExplain}}
\def\thefootnote{\arabic{footnote}}

\clearpage
\section{Introduction}
\label{sec:intro}
Knowledge Distillation (KD)~\citeMain{kd} has proven 
effective 
for improving classification accuracies of relatively small `student' models, by training them to match the logit distribution of larger, more powerful `teacher' models. Despite its simplicity, this approach can be sufficient for the students to match the teacher’s accuracy, while requiring only a fraction of the computational resources of the teacher~\citeMain{consistency}.
Recent findings, however, show that
while the students might match the teacher’s {accuracy},
the knowledge is nonetheless not distilled faithfully \citeMain{fidelity}.

Faithful KD, \ie a distillation that ensures that the teacher's and the student's functions share properties beyond classification accuracy, is however desirable for many reasons.
\Eg, the lack of {model} agreement %
\citeMain{fidelity}
{can}
hurt the user experience when updating machine-learning-based applications \citeMain{bansal2019updates,yan2021positive}. {Similarly}, if the students use different \features than the teachers, they might not be \emph{right for the right reasons} \citeMain{rrr}. {Further}, given the recent AI Act proposal by European legislators \citeMain{eu2022aiact}, it is likely that model interpretability % 
will play an increasingly important role and become an intrinsic part of the model functionality.
{To maintain the \emph{full} functionality of a model, KD should thus ensure that the students allow for the same degree of model interpretability as the teachers.}

{To address this, in this work 
{we discuss three desiderata for faithful KD and study}
if promoting explanation similarity using commonly used model explanations such as \gradcam \citeMain{gradcam} or those of the recently proposed \bcos models \citeMain{bcos} can increase 
{the faithfulness of distillation}. This should be the case if such explanations indeed reflect meaningful aspects of the models' `internal reasoning'}. Concretely, we propose \textbf{`explanation-enhanced' KD (\eKD)}, 
a simple, parameter-free, and model-agnostic addition to KD in which we train the student to also match the teacher's explanations.

Despite its simplicity, \eKD significantly advances towards faithful distillation in a variety of settings (\cref{fig:final-teaser}). Specifically, {\eKD improves student accuracy, ensures that the students learn to be right for the right reasons, and inherently promotes consistent explanations between teachers and students.} Moreover, the benefits of \eKD are robust to limited data, approximate explanations, and across model architectures.
In short, we make the \textbf{following contributions}:
\begin{enumerate}[wide, label={\textbf{(\arabic*)}}, itemsep=0em,  labelwidth=0em, labelindent=0pt, topsep=0em]
\item We propose \textbf{{explanation-enhanced KD} (\eKD)} and train the students to not only match the teachers' logits, but also their explanations (\cref{sec:exploss}); {for this, we use \bcos and \gradcam explanations.}
 This not only yields competitive students in terms of accuracy, but also significantly improves KD faithfulness on the ImageNet \citeMain{imagenet}, \waterbirds \citeMain{wb100,gals}, and \voc \citeMain{voc} datasets.
\item We discuss three desiderata for measuring the faithfulness of KD (\cref{sec:exploss:settings}). We evaluate whether the student is performant and has high agreement with the teacher (\desione), examine whether students learn to use the same \features as a teacher that was guided to be `right for the right reasons' even when distilling with biased data (\desitwo), and explore whether they learn the same explanations and architectural priors as the \mbox{teacher (\desithree)}.
\item We show \eKD to be a robust approach for improving knowledge distillation, which provides consistent gains across model architectures {and} with limited \mbox{data. Further}, \eKD is even robust to using cheaper `approximate' explanations. Specifically, for this we propose `\textbf{frozen explanations}' which are only computed once and, during training, undergo the same augmentations as images (\cref{sec:method:robustness}). 
\end{enumerate} 
\section{Related Work}
\label{sec:related-work}
\myparagraph{Knowledge Distillation (KD)} has been introduced to compress larger {models into} more efficient models for cost-effective deployment \citeMain{kd}. Various approaches have since been proposed, which we group into three types in the following {discussion}:
\emph{logit-}\citeMain{kd,dkd,consistency}, \emph{feature-}\citeMain{fitnet, attention, jacobian, reviewkd}, \mbox{and \emph{explanation-based KD}\citeMain{catkd,xdistillation,TuberculosisGcam}}. 

\underline{\smash{\emph{Logit-based KD}}} \citeMain{kd}, which optimizes the logit distributions of teacher and student to be similar, can suffice to match their accuracies, as long as the models are trained for long enough (`patient teaching') and the models' logits are based on the same images (`consistent teaching'), see \citeMain{consistency}. However, \citeMain{fidelity} showed that despite such a careful setup, the function learnt by the student can still significantly differ from the teacher's by comparing the agreement between the two. We expand on \citeMain{fidelity} and introduce additional settings to assess
{the faithfulness of distillation}, and show that it can be significantly improved by a surprisingly simple explanation-matching approach. While \citeMain{ojha2022knowledge} finds that KD does seem to transfer additional properties to the student, by showing that \gradcam explanations of the students are more similar to the teacher's than those of an independently trained model, we show that explicitly optimizing for explanation similarity  significantly improves this {w.r.t.~logit-based KD}, whilst also yielding important additional benefits such as higher robustness to distribution shifts.

\underline{\smash{\emph{Feature-based KD}}} approaches \citeMain{fitnet, attention, jacobian, reviewkd, NORM} provide additional information to the students by optimizing some of the students' intermediate activation maps to be similar to those of the teacher. For this, specific choices regarding which layers of teachers and students to match need to be made and these approaches are thus architecture-dependent. In contrast, our proposed \eKD is architecture-agnostic as it matches only the explanations of the models' predictions.

\underline{\smash{\emph{Explanation-based KD}}} approaches have only recently begun to emerge \citeMain{catkd,xdistillation,TuberculosisGcam} and these are conceptually most related to our work. In \catkd \citeMain{catkd}, the authors match class activation maps (CAM \citeMain{cam}) of students and teachers. As such, \catkd can also be considered an `explanation-enhanced' KD (\eKD) approach.  However, the explanation aspect of the CAMs plays only a secondary role in \citeMain{catkd}, as the authors even reduce the resolution of the CAMs to 2$\times$2 and faithfulness is not considered. In contrast, we explicitly introduce
\eKD to promote faithful distillation and evaluate faithfulness across multiple settings. Further, similar to our work, \citeMain{xdistillation} argues that explanations can form part of the model functionality and should be considered in KD. For this, the authors train an additional autoencoder to mimic the explanations of the teacher; explanations and predictions are thus produced by separate models. In contrast, we optimize the students directly to yield similar explanations as the teachers in a simple and parameter-free manner.

\myparagraph{Fixed Teaching.}{
\citeMain{relabel, fastkd, faghri2023reinforce} explore pre-computing the logits at the start of training to limit the computational costs due to the teacher. In addition to pre-computing \textit{logits}, we pre-compute \textit{explanations} and show how they can nonetheless be used to guide the student model during distillation.

\myparagraph{Explanation Methods.} 
To better understand the decision making process of DNNs, many explanation methods have been proposed in recent years~\citeMain{gradcam,ribeiro2016should,bach2015pixel,bcos}. For our \eKD experiments, we take advantage of the differentiability of attribution-based explanations and train the student models to yield similar explanations as the teachers. In particular, we evaluate both a popular post-hoc explanation method (\gradcam \citeMain{gradcam}) as well as the model-inherent explanations of the recently proposed B-cos models \citeMain{bcos,bcosv2}. 

\myparagraph{Model Guidance.} \eKD is inspired by recent advances in 
model guidance \citeMain{rrr,gao2022aligning,gao2022res,gals,modelguidance}, 
where models are guided to focus on desired \features via human annotations. Analogously, we also guide the focus of student models, but using knowledge (explanations) of a teacher model instead of a human annotator. As such, no explicit guidance annotations are required in our approach. Further, in contrast to the discrete annotations typically used in model guidance (\eg bounding boxes or segmentation masks), we use the real-valued explanations as given by the teacher model.
Our approach thus shares additional similarities with \citeMain{gals}, in which a model is guided via the attention maps of a vision-language model. Similar to our work, the authors show that this can guide the students to focus on the `right' \features. We extend such guidance to KD and discuss the benefits that this yields for {faithful distillation}. %KD fidelity.
\section{Explanation-Enhanced KD and {Evaluating Faithfulness}}
\label{sec:method}

To {promote faithful KD}, we introduce our proposed \emph{explanation-enhanced KD} (\eKD) in \cref{sec:exploss}. Then, in \cref{sec:exploss:settings}, we present three desiderata that faithful KD should fulfill and why we expect \eKD to be beneficial in the presented settings. %
Finally, in \cref{sec:method:robustness}, we describe how to take advantage of \eKD even without querying the teacher more than once per image when training the student.

\myparagraph{Notation.} 
For model $M$ and input $x$, we denote the predicted class probabilities by $p_M(x)$, obtained using softmax $\sigma (.)$ over output logits ${z_M(x)}$, possibly scaled by temperature $\tau$. We denote the class with highest probability by $\hat y_M$. 

\subsection{Explanation-Enhanced Knowledge Distillation}
\label{sec:exploss}
The logit-based knowledge distillation loss $\mathcal L_\mathit{KD}$ which minimizes KL-Divergence $\KL$ between teacher $T$ and student $S$ output probabilities is given by
\begin{align}\label{eq:vanillakd}
     \mathcal L_\mathit{KD} = \tau^{2} \KL(p_{T}(x;\tau) || p_{S}(x;\tau)) 
     = {-\tau^2}\sum_{j=1}^{c} \sigma_{j}\left(\dfrac{z_{T}}{\tau}\right) \log\sigma_{j}\left(\dfrac{z_{S}}{\tau}\right).
\end{align}
We propose to leverage advances in model explanations and explicitly include a term $\mathcal{L}_\mathit{exp}$ that promotes explanation similarity {for a more faithful distillation}:
\begin{align}
    \mathcal L = \mathcal L_\mathit{KD} + \lambda \mathcal L_\mathit{exp}.
\end{align}
Specifically, we maximize the similarity between the models' explanations, for the class $\hat y_{T}$ predicted by the teacher:
\begin{align}
\label{eq:ourloss}
\mathcal{L}_\mathit{exp} = 1-\text{sim}\left(E(T, x, \hat y_{T}), E(S, x, \hat y_{T})\right)\;.
\end{align}
Here, $E(M, x, \hat y_{T})$ denotes an explanation of model $M$ for class $\hat y_{T}$ and $\text{sim}$ a similarity function; in particular, we rely on well-established explanation methods (\eg \gradcam\xspace\citeMain{gradcam}) and use % 
cosine similarity in our experiments.\\

\myparagraph{\eKD is model-agnostic.} Note that by computing the loss only across model outputs and explanations, \eKD does not make any reference to architecture-specific details. In contrast to feature distillation approaches, which match specific blocks between teacher and student, \eKD thus holds the potential to seamlessly work across different architectures without any need for adaptation. As we show in \cref{sec:results}, this indeed seems to be the case, with \eKD improving {the distillation faithfulness} out of the box for a variety of model architectures, such as CNNs, B-cos CNNs, and even \bcos ViTs \citeMain{bcosv2}.

\subsection{Evaluating Benefits of \eKD}
\label{sec:exploss:settings}

In this section, we discuss three desiderata that faithful KD should fulfill and why we expect \eKD to be beneficial.
{While distillation methods are often compared in terms of accuracy,
our findings (\cref{sec:results}) suggest that one should also consider the following desiderata to judge a distillation method on its faithfulness.
}

\subsubsection{\desicolor{Desideratum 1:} High Agreement with Teacher.}
\label{sec:method:limited}
First and foremost, faithful KD should ensure that the student classifies any given sample in the same way as the teacher, \ie, the student should have high agreement \citeMain{fidelity} with the teacher. For inputs $\{x_i\}_{i=1}^N$ this is defined as:
\begin{align}
    \text{Agreement}(T, S) = \frac{1}{N}\sum_{i=1}^N\mathbbm{1}_{\hat{y}_{i,T}=\hat{y}_{i,S}}\;.
\end{align}
While \citeMain{fidelity} found that more data points can improve the agreement, in practice, the original dataset that was used to train the teacher might be proprietary or prohibitively large (\eg \citeMain{radford2021learning}). It can thus be desirable to effectively distill knowledge \emph{efficiently} with less data. To assess the effectiveness of a given KD approach in such a setting, we propose to use a teacher trained on a large dataset (\eg \imagenet \citeMain{imagenet}) and distill its knowledge to a student using as few as 50 images per class ($\approx$ 4\% of the data) or even {on images} of an unrelated dataset.

Compared to standard supervised training, it has been argued that KD improves the student performance by providing more information (full logit distribution instead of binary labels). Similarly, by additionally providing the teachers' explanations, we show that \eKD boosts the performance even further, especially when fewer data is available to learn the same function as the teacher (\cref{sec:result:imagenet}).
\clearpage
\subsubsection{\desicolor{Desideratum 2:} Learning the `Right' Features.}
\label{sec:method:biased}
Despite achieving high accuracy, models often rely on spurious input features (are not ``right for the right reasons'' \citeMain{rrr}), %
and can generalize better if guided to use the `right' features via human annotations. This is particularly useful in the presence of distribution shifts \citeMain{wb100}. Hence, faithful distillation should ensure that student models also learn to use these `right' features from a teacher that uses them.

To assess this, we use a binary classification dataset \citeMain{wb100} in which the background is highly correlated with the class label in the training set, %
making it challenging for models to learn to use the actual class features for classification. We use a teacher that has explicitly been guided to focus on the actual class features and to ignore the background. Then, we evaluate the student's accuracy and agreement with the teacher under distribution shift, {\ie, at test time, we evaluate on images in which the class-background correlation is reversed}. By providing additional spatial clues from the teachers' explanations to the students, we find that \eKD significantly improves performance over KD (\cref{sec:result:biased}).

\subsubsection{\desicolor{Desideratum 3:} Maintaining Interpretability.}
\label{sec:method:interpretability}
Note that the teachers might be trained to exhibit certain desirable properties in their explanations \citeMain{modelguidance}, or do so as a result of a particular training paradigm \citeMain{dino} %
or the model architecture \citeMain{bcosv2}. %

We propose two settings to test if such properties are transferred. First, we measure how well the students' explanations reflect properties the teachers were explicitly trained for, \ie how well they localize class-specific \features when using a teacher that has explicitly been guided to do so \citeMain{modelguidance}. We find \eKD to lend itself well to maintaining the interpretability of the teacher, as the explanations of students are explicitly optimized for this (\cref{sec:results:voc} \resvoc).

Secondly, we {perform a case study to} assess whether KD can transfer priors that are not \textit{learnt}, but rather inherent {to} the model architecture. 
Specifically, the explanations of \bcos ViTs have been shown to be sensitive to image shifts \citeMain{bcosv2}, even when shifting by just a few pixels. To mitigate this, the authors of \citeMain{bcosv2} proposed to use a short convolutional stem. Interestingly, in \cref{sec:result:priors} \resvit, we find that by learning from a CNN teacher under \eKD, the explanations of a ViT student without convolutions also become largely {equivariant} to image shifts, and exhibit similar patterns as the teacher. 

\subsection{\eKD with `Frozen' Explanations}
\label{sec:method:robustness}
{Especially in the `consistent teaching' setup of \citeMain{consistency}}, KD requires querying the teacher  for every training step, as the input images are repeatedly augmented. To reduce the computational cost incurred by evaluating the teacher, {recent work explores using} a `fixed teacher' \citeMain{relabel, fastkd, faghri2023reinforce}, where logits are pre-computed once at the start of training and used for all augmentations.

Analogously, we propose to use pre-computed explanations for images in the \eKD framework. For this, we apply the same augmentations (\eg cropping or flipping) to images and the teacher's explanations during distillation.
In \cref{sec:result:fixed}, we show that \eKD is robust to such `frozen' explanations, despite the fact that they of course only approximate the teacher's explanations. %
As such, frozen explanations provide a trade-off between optimizing for explanation similarity and reducing the cost due to the teacher.

\section{Results}
\label{sec:results}

In the following, we present our results. Specifically, in \cref{sec:result:imagenet} we compare KD approaches in terms of accuracy and agreement on \imagenet as a function of the distillation dataset size. Thereafter, we present the results on learning the `right' features from biased data in \cref{sec:result:biased} and on maintaining the interpretability of the teacher models in \cref{sec:result:explanations}.
Lastly, in \cref{sec:result:fixed}, we show that \eKD can also yield significant benefits with approximate `frozen' explanations (cf.~\cref{sec:method:robustness}).

Before turning to the results, however, we first provide some general details with respect to explanation methods used for \eKD and our training setup.

\myparagraph{Explanation methods.} For \eKD, we use \gradcam \citeMain{gradcam} for standard models and \bcos explanations for \bcos models, optimizing the cosine similarity as per \cref{eq:ourloss}. For \bcos, we use the dynamic weights $\mathbf{W}(\mathbf{x})$ as explanations \citeMain{bcos}.
\myparagraph{Training details.} In general, we follow the recent KD setup from \citeMain{consistency}, which has shown significant improvements for KD; results based on the setup followed by \citeMain{attention,reviewkd,catkd} can be found in the supplement.
Unless specified otherwise, we use the AdamW optimizer \citeMain{loshchilov2018decoupled} and, following \citeMain{bcos}, do not use weight decay for \bcos models. 
We use a cosine learning rate schedule with initial warmup for 5 epochs. For the teacher-student logit loss on multi-label VOC dataset, we use the logit loss following \citeMain{yang2023multi} instead of \cref{eq:vanillakd}.
For AT~\citeMain{attention}, \catkd~\citeMain{catkd}, ReviewKD~\citeMain{reviewkd}, {and CRD~\citeMain{tian2019crd}} we follow the original implementation and use cross-entropy based on the ground truth labels instead of \cref{eq:vanillakd}; for an adaptation to \bcos models, see \cref{supp:sec:impl:bcos-adaptation}. 
For each method and setting, we report the results of the best hyperparameters (softmax temperature and the methods' loss coefficients) as obtained on a separate validation set.
Unless specified otherwise, we augment images via random horizontal flipping and random cropping with a final resize to $224\mytimes224$. 
For full details, see \cref{supp:sec:impl}.

\begin{table*}[t]

\caption{
\textbf{KD on \imagenet for standard models.} For a \resnet-34 teacher and a \resnet-18 student, we show the {accuracy} and {agreement} of various KD approaches for three different distillation dataset sizes. Across all settings, \eKD yields significant accuracy and agreement gains over logit-based KD approaches (KD \cite{kd, consistency} and CRD \cite{tian2019crd}). Similar results are also observed for \bcos models, see \cref{tbl:imnet:bcos,tbl:imnet:bcos:densenet}.
}
\centering
\begin{tabular}{l@{\hskip2pt}c@{\hskip8pt}c@{\hskip20pt}c@{\hskip8pt}c@{\hskip20pt}c@{\hskip8pt}c@{\hskip2pt}}
 \multirow{2}{*}{\footnotesize
 \shortstack[c]{\bf Standard Models\\Teacher ResNet-34 \\{Accuracy} 73.3\%}}&
\multicolumn{2}{c}{\bf 50 Shots\phantom{shift}}&\multicolumn{2}{c}{\bf 200 Shots\phantom{shift}}&\multicolumn{2}{c}{\bf Full data\phantom{s}}\\[.5em]
& Acc. & Agr. & Acc. & Agr. & Acc. & Agr. \\\midrule
 \footnotesize Baseline ResNet-18\phantom{shiftt} &  23.3  &  24.8  & 47.0  & 50.2  & 69.8  &  76.8 \\\midrule\color{black}
AT \citeMain{attention} &\color{black} 38.3  &\color{black} 41.1 &\color{black} 54.7  &\color{black} 59.0  &\color{black} 69.7  &\color{black} 74.9  \\\color{black}
ReviewKD \citeMain{reviewkd} &\color{black} 51.2 &\color{black} 55.6 &\color{black} 63.0  &\color{black} 69.0  &\color{black} 71.4  &\color{black} 80.0  \\\color{black}
\catkd \citeMain{catkd} &\color{black} 32.2  &\color{black} 34.5  &\color{black} 55.7  &\color{black} 60.7  &\color{black} 70.9  &\color{black} 78.7  \\
\midrule
KD \citeMain{kd,consistency} & 49.8  & 55.5  & 63.1  & 71.9  & \bf71.8  & 81.2  \\[0.1em]
\bf + e$^2$KD \scriptsize(GradCAM) & \bf54.9 & \bf61.7  & \bf64.1  & \bf73.2 & \bf71.8  & \bf81.6 \\
& \bf{\scriptsize{\color{forestgreen}\makebox[.75em][c]{+}\makebox[1.75em][r]{ 5.1 }}}  & \bf{\scriptsize{\color{forestgreen}\makebox[.75em][c]{+}\makebox[1.75em][r]{ 6.2 }}}  & \bf{\scriptsize{\color{forestgreen}\makebox[.75em][c]{+}\makebox[1.75em][r]{ 1.0 }}}  & \bf{\scriptsize{\color{forestgreen}\makebox[.75em][c]{+}\makebox[1.75em][r]{ 1.3 }}}  & \bf{\scriptsize{\color{forestgreen}\makebox[.75em][c]{+}\makebox[1.75em][r]{ 0.0 }}}  & \bf{\scriptsize{\color{forestgreen}\makebox[.75em][c]{+}\makebox[1.75em][r]{ 0.4 }}} \\
\midrule    
CRD \citeMain{tian2019crd} & 30.0  & 31.8  & 51.0  & 54.9  & 69.4  & 74.6  \\[0.1em]
\bf + e$^2$KD \scriptsize(GradCAM) & \bf34.7 & \bf37.1  & \bf54.1 & \bf58.7 & \bf70.5  & \bf76.5 \\
& \bf{\scriptsize{\color{forestgreen}\makebox[.75em][c]{+}\makebox[1.75em][r]{ 4.7 }}}  & \bf{\scriptsize{\color{forestgreen}\makebox[.75em][c]{+}\makebox[1.75em][r]{ 5.3 }}}  & \bf{\scriptsize{\color{forestgreen}\makebox[.75em][c]{+}\makebox[1.75em][r]{ 3.1 }}}  & \bf{\scriptsize{\color{forestgreen}\makebox[.75em][c]{+}\makebox[1.75em][r]{ 3.8 }}}  & \bf{\scriptsize{\color{forestgreen}\makebox[.75em][c]{+}\makebox[1.75em][r]{ 1.1 }}}  & \bf{\scriptsize{\color{forestgreen}\makebox[.75em][c]{+}\makebox[1.75em][r]{ 1.9 }}}
\end{tabular}

\label{tbl:imnet:normal}
\end{table*}
\begin{table*}[h!]
\centering
\caption{\textbf{KD on \imagenet for \bcos models.} For a \bcos \resnet-34 teacher and a \bcos \resnet-18 student, we show the {accuracy} and {agreement} of KD approaches for three different distillation dataset sizes. Across all settings, \eKD significantly improves accuracy and agreement over vanilla KD, whilst remaining competitive with prior work.}
\begin{tabular}{l@{\hskip2pt}c@{\hskip8pt}c@{\hskip20pt}c@{\hskip8pt}c@{\hskip20pt}c@{\hskip8pt}c@{\hskip2pt}}
\multirow{2}{*}{\footnotesize
 \shortstack[c]{\bf\bcos Models\\ Teacher ResNet-34 \\{Accuracy} 72.3\%}} &\multicolumn{2}{c}{\bf 50 Shots\phantom{shift}}&\multicolumn{2}{c}{\bf 200 Shots\phantom{shift}}&\multicolumn{2}{c}{\bf Full data\phantom{s}}\\[.5em]
& Acc. & Agr. & Acc. & Agr. & Acc. & Agr. \\\midrule
 \footnotesize Baseline ResNet-18\phantom{shiftt} &  32.6  &  35.1  & 53.9  & 59.4  & 68.7  &  76.9 \\\midrule\color{black}
\color{black}AT \citeMain{attention} & \color{black}41.9   &\color{black} 45.6   &\color{black} 57.2   &\color{black} 63.7    &\color{black} 69.0   &\color{black} 77.2   \\
\color{black}ReviewKD \citeMain{reviewkd} &\color{black}  47.5   &\color{black}  53.2   &\color{black} 54.1   &\color{black} 60.8   &\color{black} 57.0    &\color{black} 64.6   \\
\color{black}\catkd \citeMain{catkd} &\color{black}  53.1   &\color{black}  59.8   &\color{black} 58.6   &\color{black} 66.4   &\color{black} 63.9   &\color{black} 73.7   \\
\midrule
KD \citeMain{kd,consistency} & 35.3   & 38.4   & 56.5   & 62.9   & 70.3   & 79.9   \\[0.1em]
\bf + e$^2$KD \scriptsize(\bcos) & \bf43.9  &\bf48.4  & \bf58.8  & \bf66.0  & \bf70.6  & \bf80.3 \\
& \bf{\scriptsize{\color{forestgreen}\makebox[.75em][c]{+}\makebox[1.75em][r]{ 8.6 }}}  & \bf{\scriptsize{\color{forestgreen}\makebox[.75em][c]{+}\makebox[1.75em][r]{10.0}}}  & \bf{\scriptsize{\color{forestgreen}\makebox[.75em][c]{+}\makebox[1.75em][r]{ 2.3 }}}  & \bf{\scriptsize{\color{forestgreen}\makebox[.75em][c]{+}\makebox[1.75em][r]{ 3.1 }}}  & \bf{\scriptsize{\color{forestgreen}\makebox[.75em][c]{+}\makebox[1.75em][r]{ 0.3 }}}  & \bf{\scriptsize{\color{forestgreen}\makebox[.75em][c]{+}\makebox[1.75em][r]{ 0.4 }}}
\end{tabular}
\label{tbl:imnet:bcos}
\end{table*}
\begin{table*}[t!]
\centering
\caption{\textbf{KD and `frozen' KD (\snowflake) on \imagenet for \bcos models for a  DenseNet-169 teacher.} Similar to the results in \cref{tbl:imnet:bcos}, we find that \eKD adds significant gains to `vanilla' KD across dataset sizes (50 Shots, 200 Shots, full data) and, as it does not rely on matching specific blocks between architectures (cf.~\citeMain{reviewkd,attention}), it seamlessly works across architectures. Further, \eKD can also be used with `frozen' (\snowflake) explanations by augmenting images and pre-computed explanations jointly (\cref{sec:method:robustness}).}
\begin{tabular}{l@{\hskip2pt}c@{\hskip8pt}c@{\hskip20pt}c@{\hskip8pt}c@{\hskip20pt}c@{\hskip8pt}c@{\hskip2pt}} \multirow{2}{*}{\footnotesize
 \shortstack[c]{\bf \bcos Models\\Teacher DenseNet-169 \\{Accuracy} 75.2\%}} &\multicolumn{2}{c}{\bf 50 Shots\phantom{shift}}&\multicolumn{2}{c}{\bf 200 Shots\phantom{shift}}&\multicolumn{2}{c}{\bf Full data\phantom{s}}\\[.5em]
& Acc. & Agr. & Acc. & Agr. & Acc. & Agr. \\\midrule
 \footnotesize Baseline ResNet-18\phantom{shiftt} &  32.6  &  34.5  & 53.9  & 58.4  & 68.7  &  75.5 \\\midrule
KD \citeMain{kd,consistency} & 37.3   & 40.2   & 51.3   & 55.6   & 71.2   & 78.8   \\[0.1em]
\bf + e$^2$KD \scriptsize(\bcos) & \bf45.4 & \bf49.0  & \bf55.7 & \bf60.7 & \bf71.9 & \bf79.8 \\
& \bf{\scriptsize{\color{forestgreen}\makebox[.75em][c]{+}\makebox[1.75em][r]{ 8.1 }}}  & \bf{\scriptsize{\color{forestgreen}\makebox[.75em][c]{+}\makebox[1.75em][r]{ 8.8 }}}  & \bf{\scriptsize{\color{forestgreen}\makebox[.75em][c]{+}\makebox[1.75em][r]{ 4.4 }}}  & \bf{\scriptsize{\color{forestgreen}\makebox[.75em][c]{+}\makebox[1.75em][r]{ 5.1 }}}  & \bf{\scriptsize{\color{forestgreen}\makebox[.75em][c]{+}\makebox[1.75em][r]{ 0.7 }}}  & \bf{\scriptsize{\color{forestgreen}\makebox[.75em][c]{+}\makebox[1.75em][r]{ 1.0 }}}\\

\midrule
\snowflake KD \scriptsize  & 33.4   & 35.7   & 50.4   & 54.5   & 68.7   & 75.2   \\
\snowflake \bf + e$^2$KD \scriptsize(\bcos) & \bf38.7 & \bf41.7 & \bf53.6 & \bf58.3  & \bf69.5 & \bf76.4 \\
& \bf{\scriptsize{\color{forestgreen}\makebox[.75em][c]{+}\makebox[1.75em][r]{ 5.3 }}}  & \bf {\scriptsize{\color{forestgreen}\makebox[.75em][c]{+}\makebox[1.75em][r]{ 6.0 }}}  & \bf{\scriptsize{\color{forestgreen}\makebox[.75em][c]{+}\makebox[1.75em][r]{ 3.2 }}}  & \bf{\scriptsize{\color{forestgreen}\makebox[.75em][c]{+}\makebox[1.75em][r]{ 3.8 }}}  & \bf{\scriptsize{\color{forestgreen}\makebox[.75em][c]{+}\makebox[1.75em][r]{ 0.8 }}}  & \bf{\scriptsize{\color{forestgreen}\makebox[.75em][c]{+}\makebox[1.75em][r]{ 1.2 }}}
\end{tabular}
\label{tbl:imnet:bcos:densenet}
\end{table*}
\subsection{\eKD Improves Learning from Limited Data}
\label{sec:result:imagenet}
\myparagraph{Setup.} 
To test the robustness of \eKD with respect to the dataset size {(\cref{sec:method:limited}, \desione)}, we distill with 50 ($\approx$ 4\%) or 200 ($\approx$ 16\%) shots per class, and the full \imagenet training data; further, we also distill without access to \imagenet, performing KD on SUN397 \citeMain{sun397}, whilst still evaluating on \imagenet (and vice versa). We distill \resnet-34 \citeMain{he2016deep} teachers to \resnet-18 students for standard and \bcos models (\cref{tbl:imnet:normal,tbl:imnet:bcos}); additionally, we use a \bcos DenseNet-169 \citeMain{huang2017densely} teacher (\cref{tbl:imnet:bcos:densenet}) to evaluate distillation across architectures. For reference, we also provide results we obtained via AT~\cite{attention}, \catkd~\cite{catkd}, and ReviewKD~\cite{reviewkd}.

\myparagraph{Results.}
In \cref{tbl:imnet:normal,tbl:imnet:bcos,tbl:imnet:bcos:densenet}, we 
show that \eKD can significantly improve logit-based KD in terms of top-1 accuracy as well as top-1 teacher-agreement on \imagenet. We observe particularly large gains for small distillation dataset sizes. \Eg, for KD, accuracy and agreement for conventional (and B-cos)
models on 50 shots improve by 5.1 (\bcos: 8.6) and 6.2 (\bcos: 10.0) p.p. respectively. As \eKD is model-agnostic, we found consistent trends with another teacher (\cf \cref{tbl:imnet:bcos:densenet}), and further find it to generalise also to % 
other distillation methods (\cref{tbl:imnet:normal}; CRD).

In \cref{tbl:sun2im} {(right)}, we show that \eKD also provides significant gains when using unrelated data \citeMain{consistency}, improving student's ImageNet accuracy and agreement by 4.9 and 5.4 p.p.\ respectively, despite computing the explanations on images of SUN~\cite{sun397} dataset (\ie SUN$\rightarrow$\imagenet). Similar gains can be observed when using \imagenet images to distill a teacher trained on SUN (\ie ImageNet$\rightarrow$SUN).

\subsection{\eKD Improves Learning the `Right' Features}
\label{sec:result:biased}
\myparagraph{Setup.} 
To assess whether the students learn to use the same \features as the teacher (\cref{sec:method:biased} \desitwo), we use the \waterbirds dataset \citeMain{wb100}, a binary classification task between land- and waterbirds, in which birds are highly correlated with the image backgrounds during training. 
As teachers, we use pre-trained \resnet-50 models from \citeMain{modelguidance}, which were guided to use the bird features instead of the background; as in \cref{sec:result:imagenet}, we use conventional and \bcos models and provide results obtained via prior work for reference.%
We further demonstrate the model-agnostic aspect of \eKD by testing a variety of CNN architectures as students. In light of the findings by \citeMain{consistency} {that long teaching schedules and strong data augmentations help}, we explore three settings\footnote{Compared to \imagenet, the small size of the \waterbirds dataset allows for reproducing the `patient teaching' results with limited compute.}: (1) 700 epochs, (2) with add.~\mixup \citeMain{zhang2018mixup}, as well as (3) training 5x longer (`patient teaching').

\myparagraph{Results.} In \cref{fig:wb:ood}, we present our results on the \wbirdsshort for standard models (see \cref{supp:sec:quantitative:waterbirds} for \bcos models). We evaluate accuracy and student-teacher agreement of each method on object-background combinations not seen during training (\ie `Waterbird on Land' \& `Landbird on Water') to see how well the students learned from the teacher to rely on the `right' \features (\ie birds).

\begin{figure*}[t]
    \centering
    \includegraphics[width=0.985\textwidth]{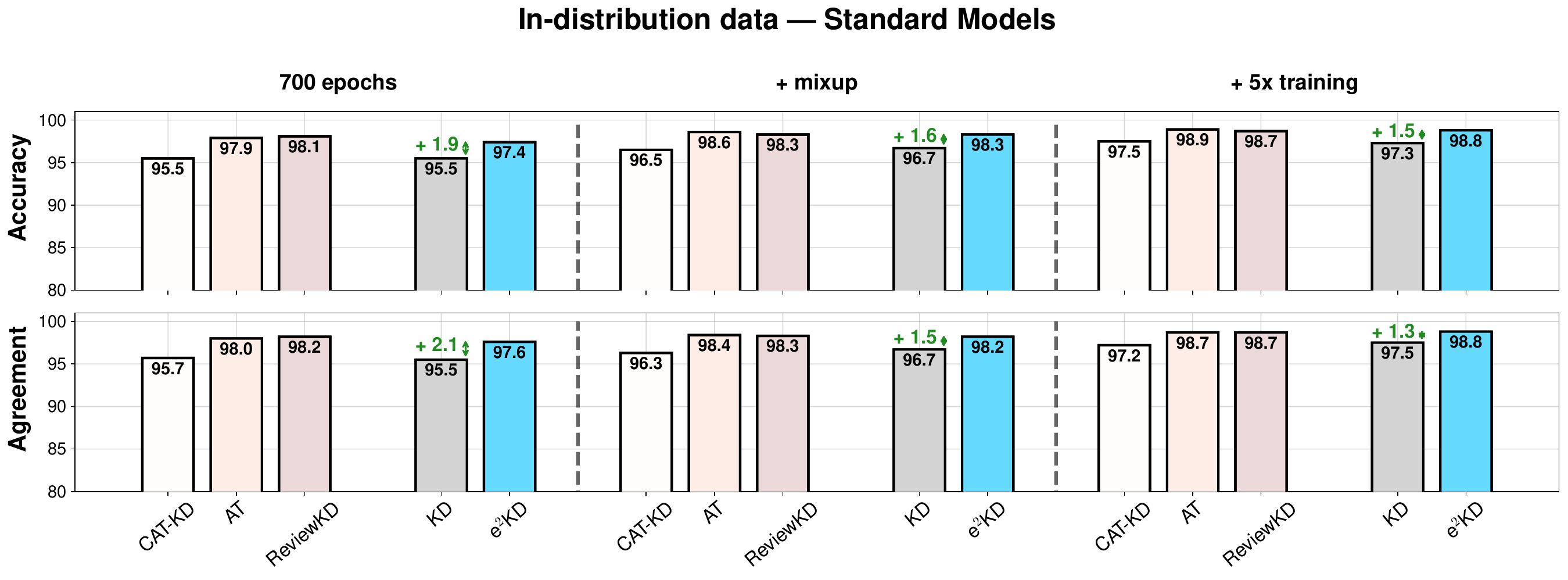}
    \includegraphics[width=0.985\textwidth]{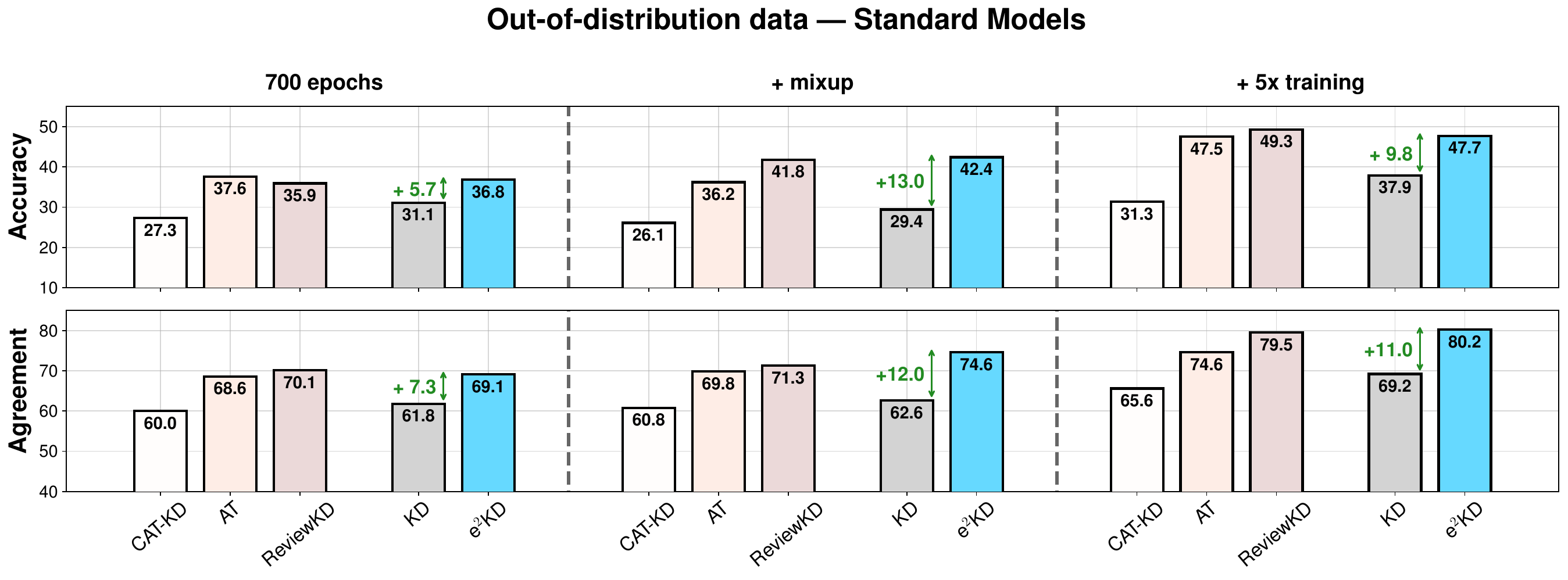}
    \caption{\textbf{KD for standard models on \waterbirds.} We show the accuracy and agreement on in-distribution ({\bf top}) and out-of-distribution ({\bf bottom}) test samples when distilling from a ResNet-50 teacher to a ResNet-18 student with various KD approaches. Following \cite{consistency}, we additionally evaluate the effectiveness of adding \mixup (\textbf{col.~2}) and, additionally, long teaching (\textbf{col.~3}). We find that our proposed \eKD provides significant benefits over vanilla KD, and is further enhanced under long teaching and \mixup. We show the performance of prior work for reference, and find that \eKD performs competitively. For results on \bcos models, see \cref{supp:sec:quantitative:waterbirds} and \cref{fig:bcos-waterbirds}.
    }
    \label{fig:wb:ood}
\end{figure*}
\begin{figure*}[h!]
    \centering
    \includegraphics[width=0.48\linewidth]{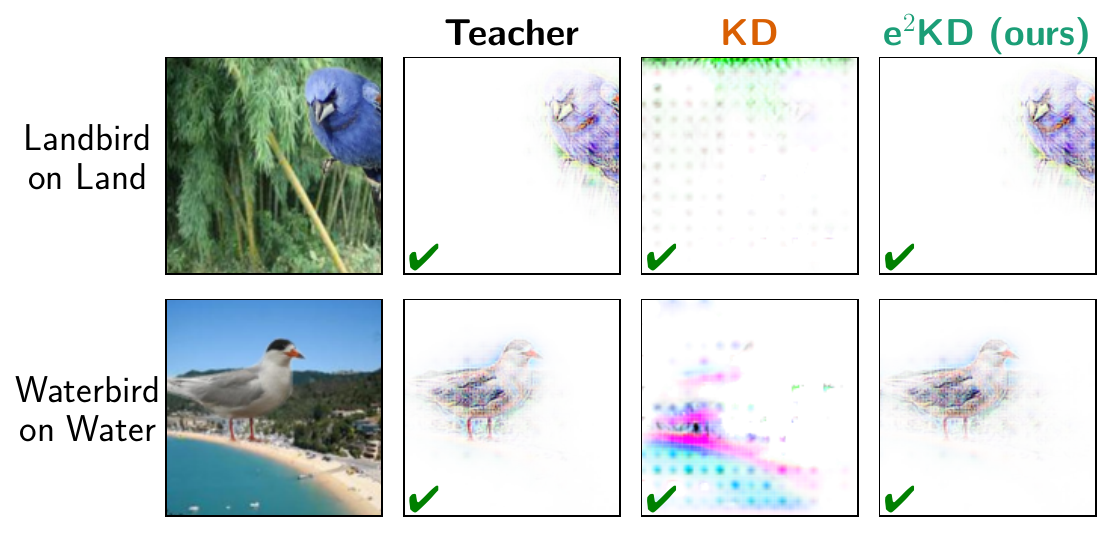} \hfill \includegraphics[width=0.48\linewidth]{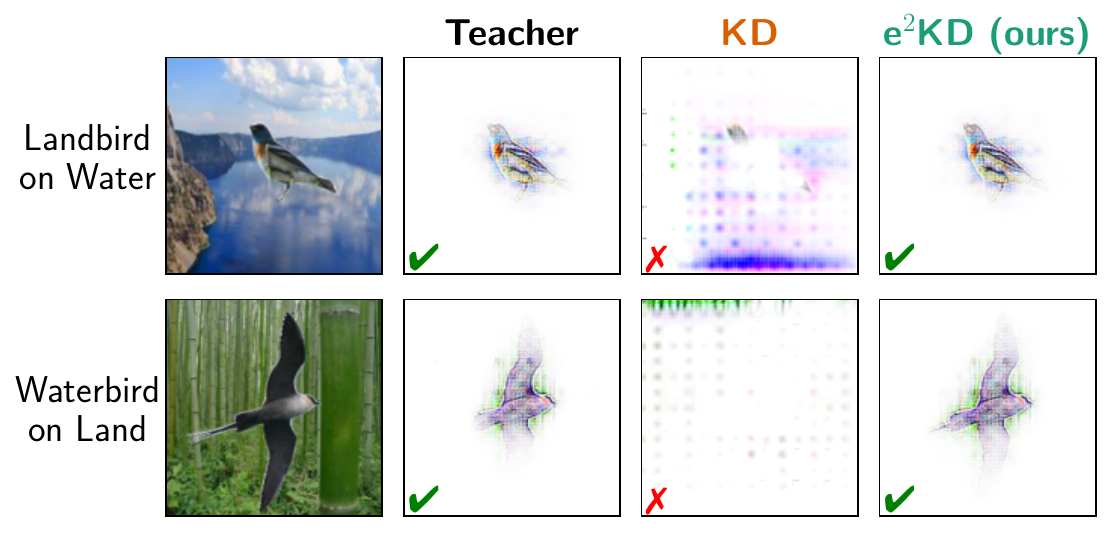}
    \caption{\textbf{Comparing explanations for KD on Waterbirds.}  
    Here we visualize \bcos explanations, when distilling a \bcos \resnet-50 teacher (\textbf{col.~2}) to a \bcos \resnet-18 student with KD (\textbf{col.~3}) and \eKD (\textbf{col.~4}). While for in-distribution data (\textbf{left}) the different focus of the models (foreground/background) does not affect the models' predictions (correct predictions marked by {\color{forestgreen} \ding{51}}), it results in wrong predictions under distribution shift (\textbf{right}, incorrect predictions marked by {\color{red}\ding{55}}). For additional qualitative results, including standard models with \gradcam explanations, see \cref{supp:sec:qualitative:waterbirds}.\label{fig:bcos-waterbirds}}
\end{figure*}

Across all settings, \eKD significantly boosts the out-of-distribution performance of KD on both accuracy and agreement. Despite its simplicity, it compares favourably to prior work, indicating that \eKD indeed promotes faithful distillation. Notably, \cref{fig:wb:ood} is also an example of how the in-distribution performance of KD methods may not fully reflect their differences.%
%For quantitative results on \bcos models, see appendix \textcolor{red}{B.2}.
We also find clear qualitative improvements in the explanations focusing on the `right' features, see \cref{fig:bcos-waterbirds} for \bcos models and \cref{supp:sec:qualitative:waterbirds} for standard models.

\clearpage
Further, consistent with \citeMain{consistency}, we find \mixup augmentation and longer training schedules to also significantly improve agreement. This provides additional evidence for the hypothesis put forward by \citeMain{consistency} that KD \emph{could} be sufficient for function matching if performed for long enough. As such, and given the simplicity of the dataset, the low resource requirements, and a clear target (100\% agreement on unseen combinations), we believe the waterbirds dataset to constitute a great benchmark for future research towards faithful KD.

Lastly, given that \eKD does not make any reference to model architecture and simply matches the explanations on top of KD, we find that it consistently improves out-of-distribution performance across different student architectures, see \cref{tab:wb_variety}. As we discuss in the next section, the model-agnostic nature of \eKD also seamlessly allows to transfer knowledge between CNNs and ViTs.

\begin{table}[h]
    \centering
    \caption{\textbf{Out-of-distribution results on Waterbirds-100 across student architectures.} We show accuracy and agreement results on out-of-distribution samples when distilling a standard \resnet-50 teacher (similar to \cref{fig:wb:ood}) to different students. \eKD results in consistent gains across students, by simply matching the explanations.}
    \begin{tabular}{l @{\hskip10pt} c@{\hskip10pt}c @{\hskip14pt} c@{\hskip10pt}c @{\hskip14pt} c c @{\hskip14pt} c@{\hskip10pt}c}
    \multirow{2}{*}{Method} & \multicolumn{2}{c}{\scriptsize \bf ConvNext \phantom{shu}} & \multicolumn{2}{c}{\scriptsize \bf EfficientNet \phantom{shi}} & \multicolumn{2}{c}{\scriptsize \bf MobileNet \phantom{shif}} & \multicolumn{2}{c}{\scriptsize \bf ShuffleNet}\\
                                            & Acc. & Agr. & Acc. & Agr. & Acc. & Agr. & Acc. & Agr. \\
    \hline
    KD                                      & 20.5& 55.5  & 27.5& 59.0  & 22.3& 57.0  & 23.1& 57.1  \\
    \bf + e$^2$KD \scriptsize(GradCAM)      & \bf 32.2& \bf 64.4 & \bf 37.8& \bf 68.7 & \bf 36.0& \bf 68.2 & \bf 37.0& \bf 68.6
    \end{tabular}
    \label{tab:wb_variety}
\end{table}

\subsection{\eKD Improves the Student's Interpretability}
\label{sec:result:explanations}
{In this section, we present results on maintaining the teacher's interpretability (\cf\cref{sec:method:interpretability} \desithree). In particular,  we show that \eKD naturally lends itself to distilling localization properties of the teacher into the students (\cref{sec:results:voc} \resvoc) and that even architectural priors of a CNNs can be transferred to ViT students (\cref{sec:result:priors} \resvit).}
\subsubsection{Distill on VOC.}
\label{sec:results:voc}
We assess how well the focused explanations are preserved.
\myparagraph{Setup.} To assess whether the students learn to give similar explanations as the teachers,
{we distill \bcos \resnet-50 teachers into \bcos \resnet-18 students on \voc \citeMain{voc} in a multi-label classification setting.}
Specifically, we use two different teachers from \citeMain{modelguidance}: one with explanations of high EPG~\cite{Wang_2020_CVPR_Workshops} score (EPG Teacher), and one with explanations of high IoU score (IoU Teacher). To quantify the students' focus, we then measure the EPG and IoU scores of the explanations with respect to the dataset's bounding box annotations in a multi-label classification setting. As these teachers are trained explicitly to exhibit certain properties in their explanations, a \textit{faithfully distilled} student should optimally exhibit the same properties.

\begin{table*}[h!]
\centering
\caption{\textbf{(Left) \eKD results on VOC.} We compare KD and \eKD when distilling from a \bcos \resnet-50 teacher guided  \citeMain{modelguidance} to either optimize for EPG (\emph{left}) or IoU (\emph{right}). Explanations of the \eKD\  {student} better align with those of the teacher, as evidenced by significantly higher EPG (IoU) scores when distilled from the EPG (IoU) teacher. \eKD students also achieve higher accuracy (F1).\textbf{{(Right)} KD on unrelated images}. A \bcos \densenet-169 teacher model, \emph{left:} trained on the SUN~\citeMain{sun397} is distilled with \imagenet (IMN$\rightarrow$SUN), and \emph{right:} trained on \imagenet is distilled with SUN (SUN$\rightarrow$IMN). In both cases, the \bcos \resnet-18 student distilled with \eKD achieves significantly higher accuracy and agreement scores than student trained via vanilla KD.}

\begin{minipage}[h]{0.3\linewidth}
\begin{tabu}
{@{\hskip0pt}l @{\hskip6pt}c @{\hskip6pt}c @{\hskip6pt}c @{\hskip8pt}|@{\hskip8pt}c @{\hskip6pt}c @{\hskip6pt}c}
&\multicolumn{6}{c}{\textbf{KD on the \vocshort Dataset}}\\[.5em]
 & \multicolumn{3}{c}{\textbf{{EPG Teacher} \phantom{shi}}} & \multicolumn{3}{c}{\textbf{{IoU Teacher}\phantom{s}}} \\[.35em]
{}             &     \underline{\bf EPG}  &      IoU  &       F1  &      EPG  &    \underline{\bf IoU}  &      F1  \\[.35em]

{ \scriptsize Teacher}   &     75.7        &       21.3     &    72.5     &      65.0      &       49.7      &   72.8    \\
{ \scriptsize Baseline}   &     50.0        &       29.0     &    58.0     &      50.0      &       29.0     &    58.0    \\
\midrule
KD          &           60.1 &     31.6 &           60.1 &  58.9      &           35.7 &           62.7 \\\bf
+ \eKD        &  \textbf{71.1} &     24.8 &  \textbf{67.6} &  60.3      &  \textbf{45.7} &  \textbf{64.8}
\label{tbl:voc}
\end{tabu}
\end{minipage}
\hfill
\begin{minipage}[h]{0.38\linewidth}
\centering
\begin{tabu}{@{\hskip0pt}l@{\hskip6pt}c@{\hskip6pt}c@{\hskip12pt}|@{\hskip12pt}c@{\hskip6pt}c}
\multicolumn{5}{c}{\textbf{KD on Unrelated Images}}\\[.5em]
\multicolumn{3}{l}{\bf {IMN$\rightarrow$SUN}} & \multicolumn{2}{c}{\hspace{-6pt}\bf {SUN$\rightarrow$IMN}}\\[.35em]
 & Acc. & Agr. & Acc. &  Agr.\\[.35em]
&   60.5 &   -    &   75.2 & -\\
&   57.7 &   67.9 &   68.7 &  75.5\\
\midrule
&           53.5 &              65.0       &   14.9        &    16.7\\
&  \textbf{54.9} &     \textbf{67.7}   & \bf19.8       &\bf22.1
\end{tabu}
\end{minipage}

\label{tbl:sun2im}
\end{table*}
\myparagraph{Results.}
{As we show in \cref{tbl:voc}, the explanations of an \eKD student indeed more closely mirror those of the teacher than a student trained via vanilla KD:}
{\eKD students} exhibit significantly higher EPG when distilled from the EPG teacher (EPG: 71.1 vs.~60.3) and vice versa (IoU: 45.7 vs.~24.8). In contrast,  `vanilla' KD students show only minor differences (EPG: 60.1 vs.~58.9; IoU: 35.7 vs.~31.6). These improvements also show qualitatively (\cref{fig:bcos-voc}), with the \eKD students reflecting the teacher's focus much more faithfully in their explanations.

While this might be expected as \eKD explicitly optimizes for explanation similarity, we would like to highlight that this not only ensures that the desired properties of the teachers are better represented in the student model, but also significantly improves the students' performance (\eg, F1: 60.1$\rightarrow$67.6 for the EPG teacher). As such, we find \eKD to be an easy-to-use and effective addition to vanilla KD for improving both interpretability as well as task performance.%
\begin{figure*}[h!]
    \includegraphics[width=0.475\linewidth]{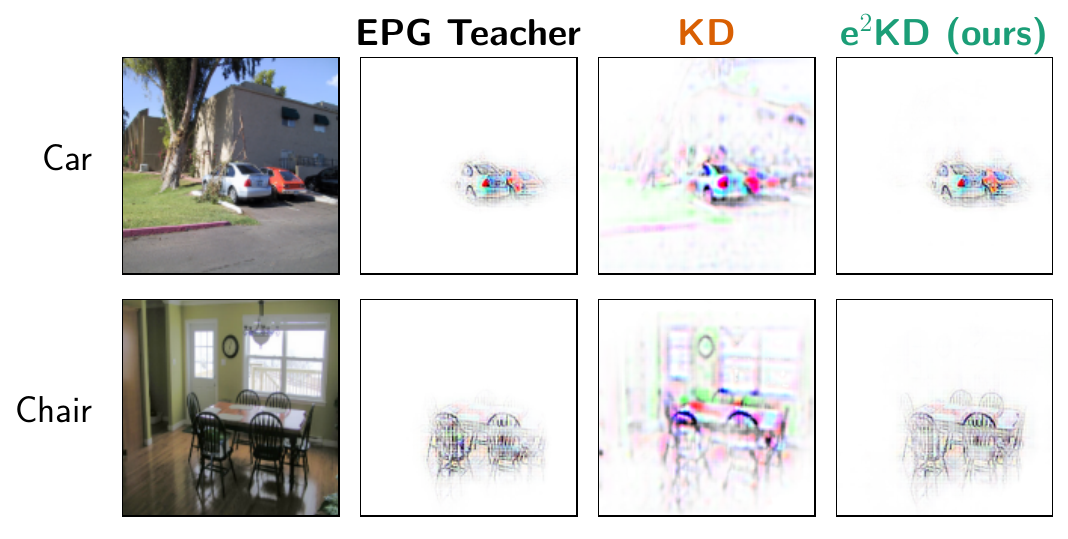}%
    \hfill%
    \includegraphics[width=0.499\linewidth]{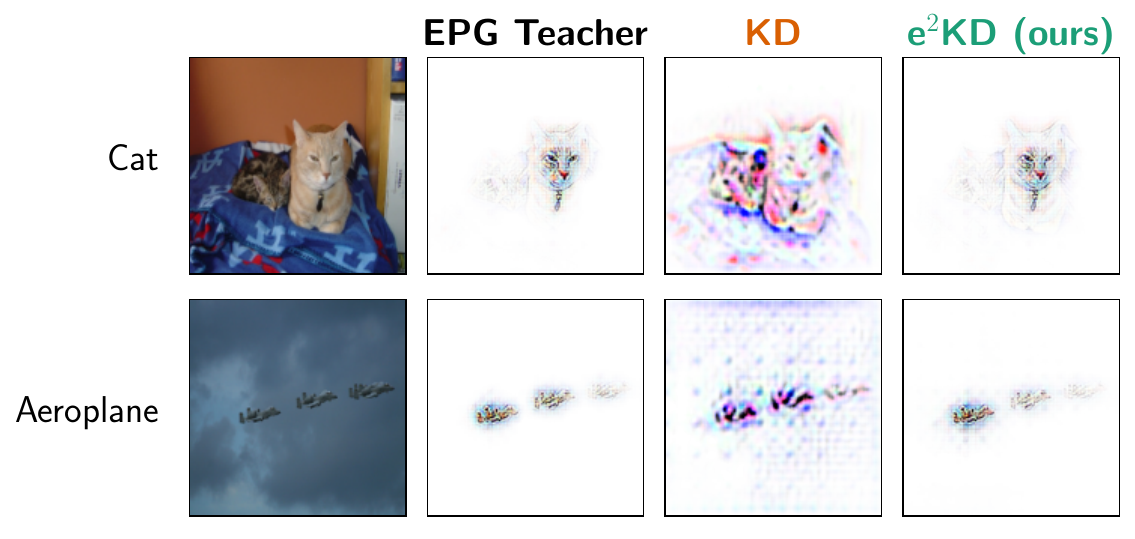}
    \caption{\textbf{Maintaining focused explanations.} We visualize \bcos explanations, when distilling a \bcos\resnet-50 teacher that has been trained to not focus on confounding input features ({\bf col.~2}), to a \bcos\resnet-18 student with KD ({\bf col.~3}) and \eKD ({\bf col.~4}).
    {Explanations of}
    \eKD students are significantly closer to the teacher's (and hence more human-aligned). Samples are drawn from the \vocshort test set, with all models correctly classifying the shown samples. For more qualitative results, see \cref{supp:sec:qualitative:voc}.
    }
    \label{fig:bcos-voc}
\end{figure*}

\subsubsection{Distill to ViT.} We assess if inductive biases of CNN can be distilled to ViT.\label{sec:result:priors}
\myparagraph{Setup.} To test whether students learn architectural priors of the models, we evaluate whether a \bcos ViT$_\text{Tiny}$ student can learn to give explanations that are similar to those of a pretrained CNN (\bcos DenseNet-169) teacher model;
for this, we again use the \imagenet dataset.

\begin{figure*}[t]
\centering
\begin{minipage}[l]{5.3cm}
    \centering
    \begin{small}
    \begin{tabular}{l@{\hskip8pt}c@{\hskip3pt}c}
    \bf \scriptsize Method&\bf \scriptsize  Acc.&\bf \scriptsize  Agr.\\[.5em]
     T: \bcos \densenet-169 &   75.2 &   - \\
     B: \bcos ViT$_{\text{Tiny}}$ &   60.0 &   64.6\\
    \hline
     KD &  64.8 &  70.1
    \\
     \bf + \eKD &  \textbf{66.3} &  \textbf{71.8}
    \end{tabular}
    \end{small}
    
    \includegraphics[width=\textwidth]{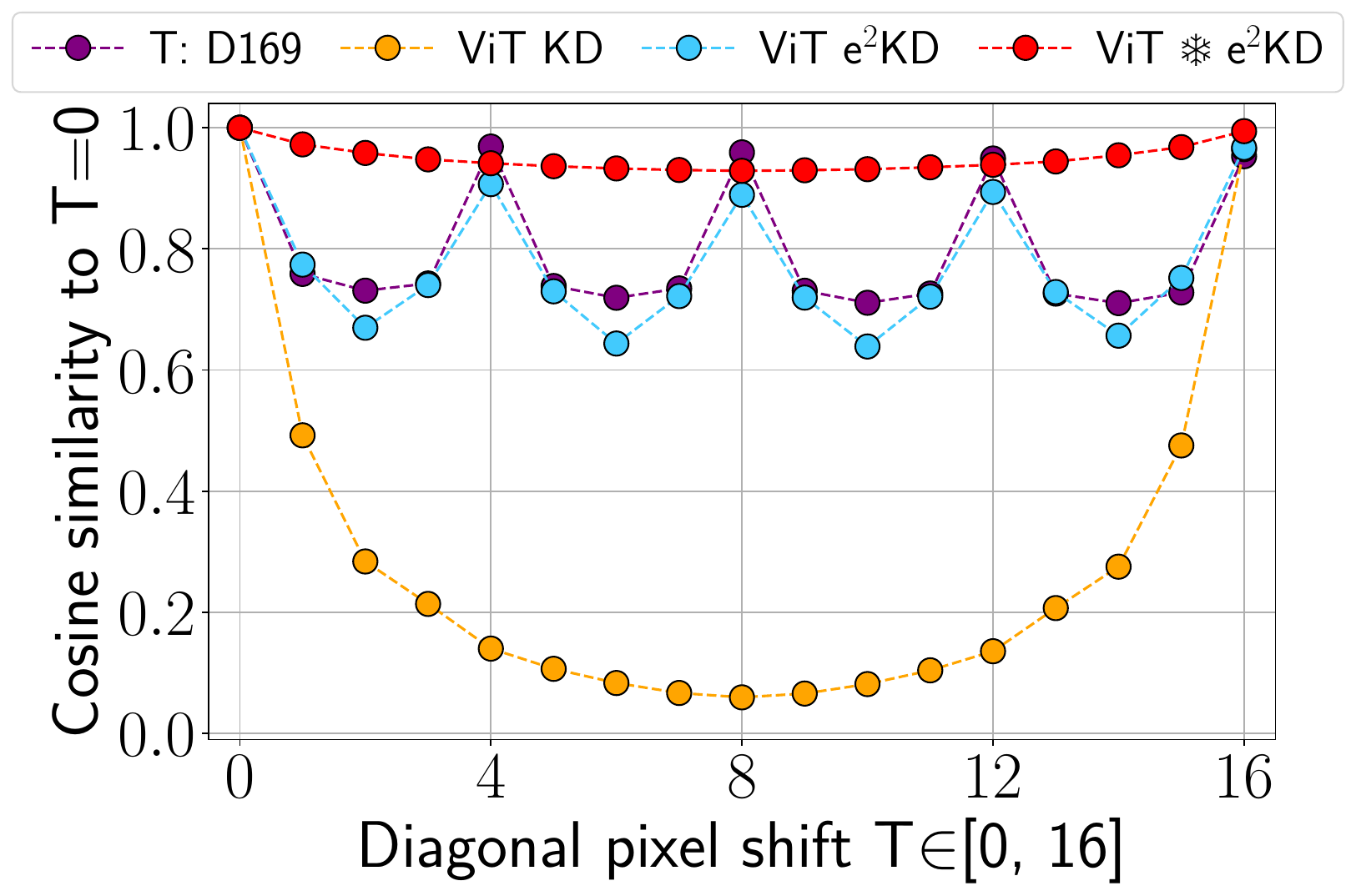}%
\end{minipage}\hspace{0.3cm}%
\begin{minipage}[l]{5.9cm}
    \centering
    \includegraphics[width=\textwidth]{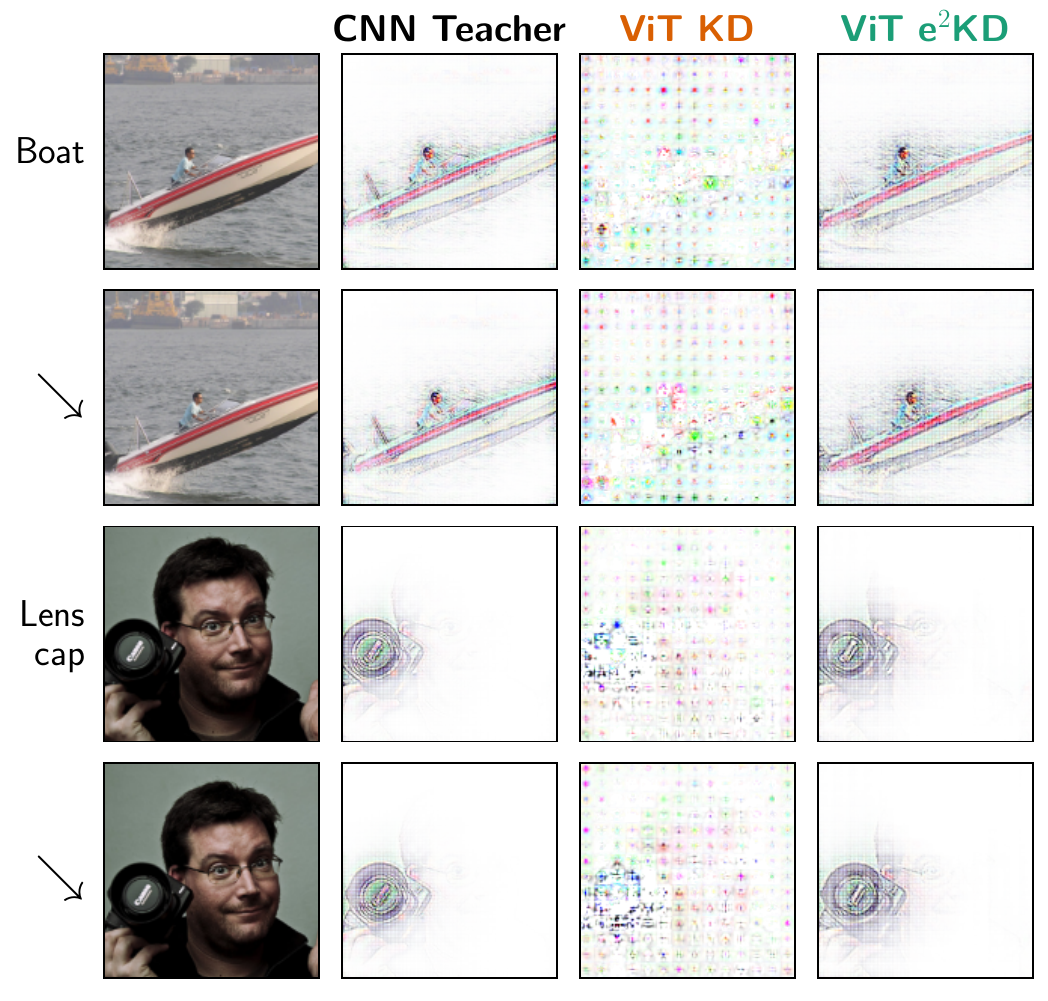}
\end{minipage}

\centering
\caption{\textbf{Distilling inductive biases (CNN$\rightarrow$ViT).} We distill a \bcos \densenet-169 teacher to a \bcos ViT$_\text{Tiny}$. \textbf{Top-Left:} \eKD yields significant gains in accuracy and agreement. \textbf{Bottom-Left:} Cosine similarity of explanations for shifted images w.r.t.~the unshifted image ($T$=0). With \eKD (blue) the ViT student learns to mimic the shift periodicity of the teacher (purple), despite the inherent periodicity of 16 of the ViT architecture (seen for vanilla KD, yellow). Notably, \eKD with frozen explanations yields shift-equivariant students (red), see also \cref{sec:result:priors} \resvit. \textbf{Right:} \eKD significantly improves the explanations of the ViT model, thus maintaining the utility of the explanations of the CNN teacher model. While the explanations for KD change significantly under shift (subcol.~3), for \eKD (subcol.~4), as with the CNN teacher (subcol.~2), the explanations remain consistent. See also \cref{supp:sec:qualitative:vit}.
}
\label{fig:vit_results}
\label{fig:shift-equivariance}
\end{figure*}
\myparagraph{Results.}
In line with the results of the preceding sections, we find {(\cref{fig:vit_results}, left)} that \eKD significantly improves the accuracy of the ViT student model (64.8$\rightarrow$66.3), as well as the agreement with the teacher (70.1$\rightarrow$71.8).

Interestingly, we find that the ViT student's explanations seem to become similarly robust to image shifts as those of the teacher (\cref{fig:vit_results}, bottom-left and right.). Specifically, note that the image tokenization of the ViT model using vanilla KD (extracting non-overlapping patches of size 16$\times$16) induces a periodicity of 16 with respect to image shifts $T$, see, \eg, \cref{fig:vit_results} (bottom-left, yellow curve): here, we plot the cosine similarity of the explanations\footnote{We compute the similarity of the intersecting area of the explanations.} at various shifts with respect to the %
explanation given for the %
original, unshifted image ($T$=0). %
In contrast, due to smaller strides (stride$\myin\{1, 2\}$ for any layer) and overlapping convolutional kernels, the CNN teacher model is inherently more robust to image shifts, see \cref{fig:vit_results} (purple curve), exhibiting a periodicity of 4. 
A ViT student {trained} via %
\eKD %
learns to mimic the behaviour of the teacher %
(see \cref{fig:vit_results}, blue curve) %
and exhibits the same periodicity, indicating that \eKD indeed helps the student learn a function more similar to the teacher.

In \cref{fig:vit_results} (right), we see that \eKD also significantly improves the explanations of the ViT model. We show explanations for original and 8-pixel diagonally shifted ($\searrow$) images. Our ViT's explanations are more robust to shifts and more interpretable, thus maintaining the utility of the explanations of the teacher.

\subsection{\eKD with Frozen Explanations}
\label{sec:result:fixed}
In the previous sections, we showed that \eKD is a robust approach that provides consistent gains even when only limited data is available (see \cref{sec:result:imagenet}) and works across different architectures (\eg, DenseNet$\rightarrow$\resnet or DenseNet$\rightarrow$ViT, see \cref{sec:result:imagenet,sec:result:priors} \resvit).
In the following, we show that \eKD even works when only `approximate' explanations for the teacher are available \mbox{(\cf\cref{sec:method:robustness})}.

\myparagraph{Setup.}
{To test the robustness of \eKD when using frozen explanations, we distill from a \bcos DenseNet-169 teacher to a \bcos ResNet-18 student using pre-computed, frozen explanations on the \imagenet dataset. We also evaluate across varying dataset sizes, as in \cref{sec:result:imagenet}.}

\myparagraph{Results.}
\cref{tbl:imnet:bcos:densenet} (bottom) shows 
that \eKD with frozen explanations is effective for improving both the accuracy and agreement over KD with frozen logits across dataset sizes (\eg accuracy: 33.4$\rightarrow$38.7 for 50 shots). Furthermore, \eKD with frozen explanations also outperforms vanilla KD under both metrics when using limited data (\eg accuracy: 37.3$\rightarrow$38.7 for 50 shots). As such, a frozen teacher constitutes a more cost-effective alternative for obtaining the benefits of \eKD, whilst also highlighting its robustness to using `approximate' explanations. %

Our results also indicate that it might be possible to instill desired properties into a DNN model even beyond knowledge distillation. Note that the frozen explanations are \emph{by design} equivariant explanations across shifts and crops. Based on our observations for the ViTs (\cf \cref{sec:result:priors}), we thus expect a student trained on frozen explanations to become almost \emph{fully shift-equivariant}, which is indeed the case for our ViT students (see \cref{fig:vit_results}, bottom-left, red curve, ViT \snowflake\eKD).

\section{Conclusion}
\label{sec:conclusion}
We proposed a simple approach to promote the faithfulness of knowledge distillation (KD) by explicitly optimizing for the explanation similarity between the teacher and the student, and showed its effectiveness in distilling the teacher's properties under multiple settings. Specifically, \eKD helps the student (1) achieve competitive and often higher accuracy and agreement than vanilla KD, (2) learn to be `right for the right reasons', and (3) learn to give similar explanations as the teacher, \eg even when distilling from a CNN teacher to a ViT student. Finally, we showed that \eKD is robust in the presence of limited data, approximate explanations, and across model architectures.  In short, we find \eKD to be a simple but versatile addition to KD that allows for a more faithful distillation of the teacher, whilst also maintaining competitive task performance.

\clearpage
{
\bibliographystyle{splncs04}
\bibliography{main_bib}
}

\clearpage
\appendix
\renewcommand\thesection{\Alph{section}}
\numberwithin{equation}{section}
\numberwithin{figure}{section}
\numberwithin{table}{section}
\renewcommand{\thefigure}{\thesection\arabic{figure}}
\renewcommand{\thetable}{\thesection\arabic{table}}
\crefname{appendix}{Sec.}{Secs.}

{\onecolumn 
{\begin{center}
\Large\bf
\phantom{skip}\\[.25em]
{Good Teachers Explain: Explanation-Enhanced Knowledge Distillation}\\[1em]
\large
Appendix
\end{center}
}
\newcommand{\additem}[2]{%
\item[\textbf{(\ref{#1})}] 
    \textbf{#2} \dotfill\makebox{\textbf{\pageref{#1}}
    }
}

\newcommand{\myindent}{.5em}
\newcommand{\addsubitem}[2]{%
\vspace{.4em}
    \textbf{(\ref{#1})}
        \hspace{\myindent} #2 \\    
}

\newcommand{\adddescription}[1]{\vspace{.5em}
\begin{adjustwidth}{0cm}{0cm}
#1
\end{adjustwidth}
}

\noindent In this supplement to our work on explanation-enhanced knowledge distillation (\eKD), we provide:
\\[1em]

\begin{adjustwidth}{0.15cm}{0.15cm}
\begin{enumerate}[label={({\arabic*})}, topsep=1em, itemsep=.75em]
    \additem{supp:sec:qualitative}{Additional Qualitative Results}
    \adddescription{In this section, we provide additional qualitative results for each evaluation setting.
    Specifically, we show qualitative results of the model explanations of standard models (\gradcam) and \bcos models (\bcos explanations) for KD and \eKD for the following:}\ \\
    \addsubitem{supp:sec:qualitative:waterbirds}{Learning the `right' features (\waterbirds).}
    \addsubitem{supp:sec:qualitative:voc}{Maintaining focused explanations (\voc).}
    \addsubitem{supp:sec:qualitative:vit}{Distilling architectural priors (CNN$\rightarrow$ViT on \imagenet).}
    
    \additem{supp:sec:quantitative}{Additional Quantitative Results}
    \adddescription{In this section, we provide additional quantitative results:}\ \\
    \addsubitem{supp:sec:quantitative:oldrecipe}{Reproducing previously reported \imagenet results for prior work.}
    \addsubitem{supp:sec:quantitative:waterbirds}{In- and out-of-distribution results for \bcos models on \mbox{\wbirdsshort.}}
    \addsubitem{supp:sec:quantitative:cat-kd}{Detailed Comparison with respect to \catkd}
    \additem{supp:sec:impl}{Implementation Details}
    \adddescription{In this section, we provide implementation details, including the setup used in each experiment and the procedure followed to adapt prior work to B-cos models. Code will be made available on publication.}\ \\
    \addsubitem{supp:sec:impl:training}{Training details.}
    \addsubitem{supp:sec:impl:bcos-adaptation}{Adaptation of prior work to B-cos networks.}
\end{enumerate}
\end{adjustwidth}

}

\setlength{\parskip}{.5em}
\clearpage
\section{Additional Qualitative Results}
\label{supp:sec:qualitative}

\subsection{Learning the `Right' Features}
\label{supp:sec:qualitative:waterbirds}
In this section we provide qualitative results on the \waterbirds dataset \citeS{wb100S,galsS}. We show GradCAM explanations \citeApp{gradcam} for standard models and \bcos explanations for \bcos models \citeS{bcosS,bcosv2S}. In \cref{fig:wb:qual_id_supp}, we show explanations for in-distribution (\ie `Landbird on Land' and `Waterbird on Water') test samples, and in \cref{fig:wb:qual_ood_supp} we show them for out-of-distribution samples (\ie `Landbird on Water' and `Waterbird on Land'). Corresponding quantitative results can be found in \cref{supp:sec:quantitative:waterbirds}.

From \cref{fig:wb:qual_id_supp}, we observe that the explanations of the teacher and vanilla KD student may significantly differ (for both standard and \bcos models): while the teacher is focusing on the bird, the student may use spuriously correlated input-features (\ie background). %
We observe that \eKD is successfully promoting explanation similarity and keeping the student `right for right reasons'.
While in \cref{fig:wb:qual_id_supp} (\ie in-distribution data) all models correctly classify the samples despite the difference in their focus, in \cref{fig:wb:qual_ood_supp} (\ie out-of-distribution) we observe that the student trained with \eKD is able to arrive at the correct prediction, whereas the vanilla KD student wrongly classifies the samples based on the background.

\begin{figure}[h]
    \begin{subfigure}[c]{\textwidth}
    \centering
    \includegraphics[width=0.99\textwidth]{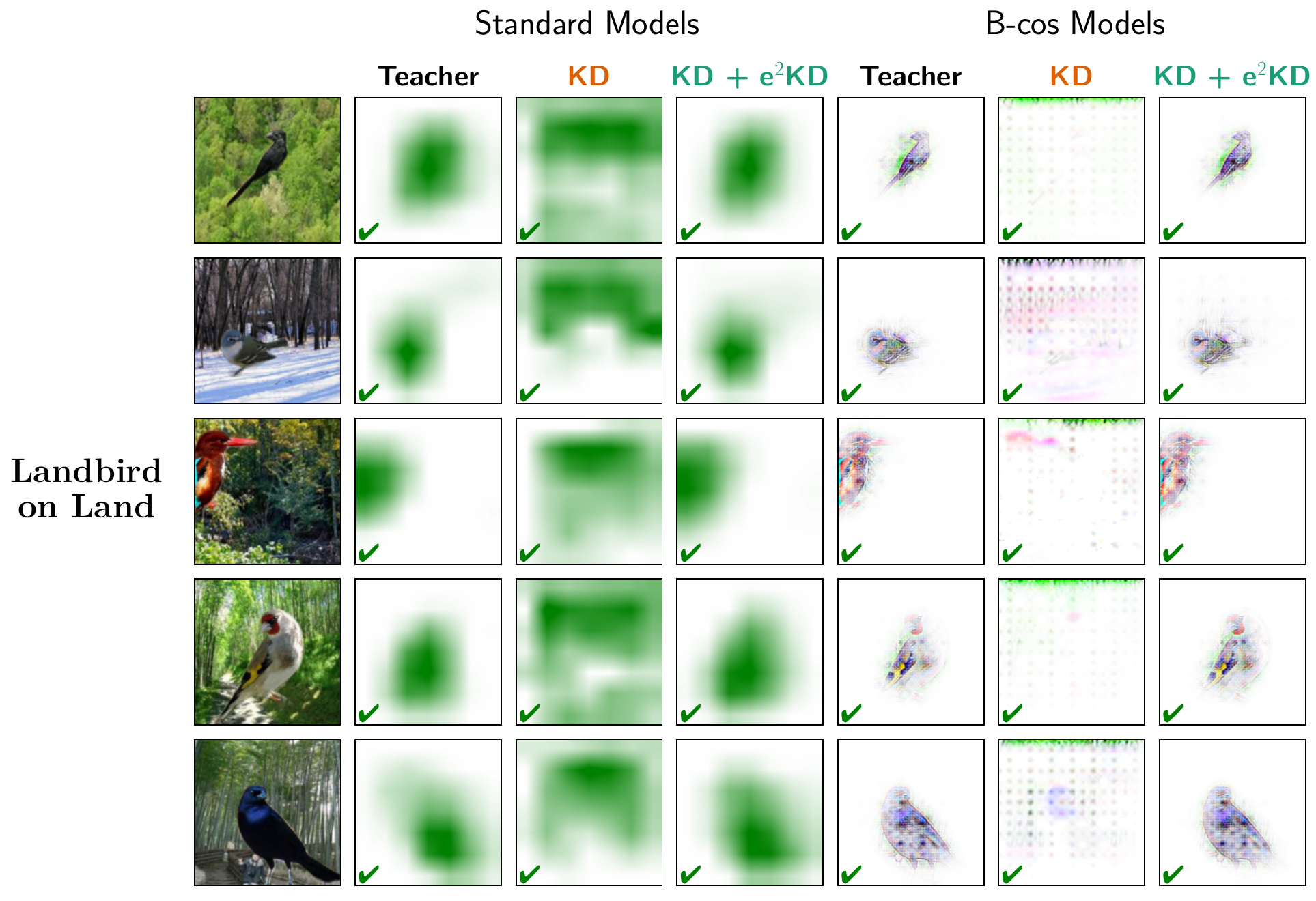}
    \end{subfigure}
    
    \vspace{1.5em}
    \begin{subfigure}[c]{\textwidth}
    \centering
    \includegraphics[width=0.99\textwidth]{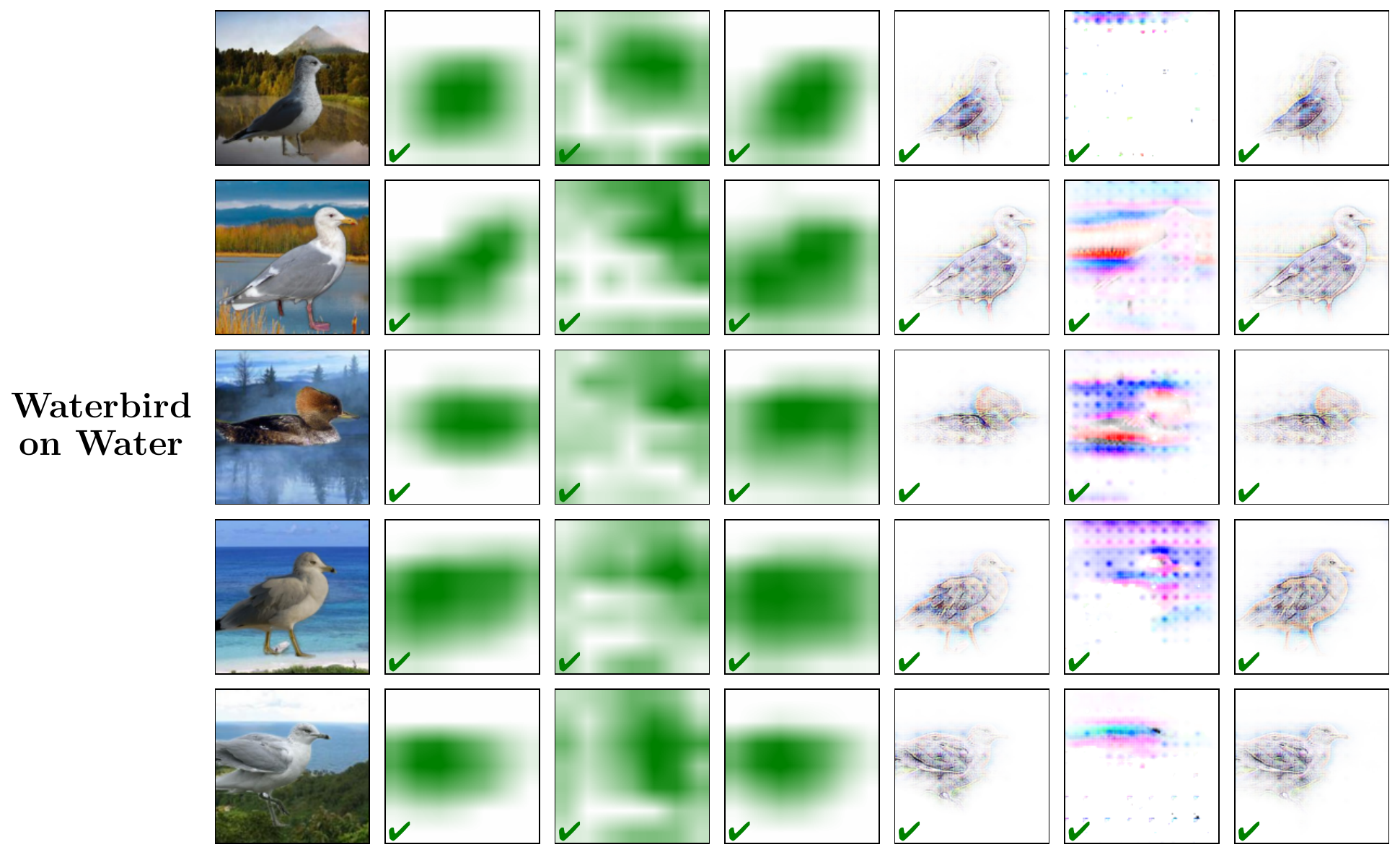}
    \end{subfigure}
    \vspace{-1em}
    \caption{\textbf{In-distribution samples for distillation on biased data using the \waterbirds dataset}. We show explanations for both standard models (cols.~2-4) and \bcos models (cols.~5-7), given both in-distribution groups: `Landbird on Land' (\textbf{top half}) and `Waterbird on Water' (\textbf{bottom half}). We find that  \eKD approach (col.~4 and 7) is effective in preserving the teacher's focus (col.~2 and 5) to the bird instead of the background as opposed to vanilla KD (col.~3 and 6). Correct and incorrect predictions marked by {\color{forestgreen}\ding{51}} and {\color{red}\ding{55}} respectively.}
    \label{fig:wb:qual_id_supp}
\end{figure}
\clearpage

\begin{figure}[h!]
    \centering
    \begin{subfigure}[c]{\textwidth}
    \centering
    \includegraphics[width=0.99\textwidth]{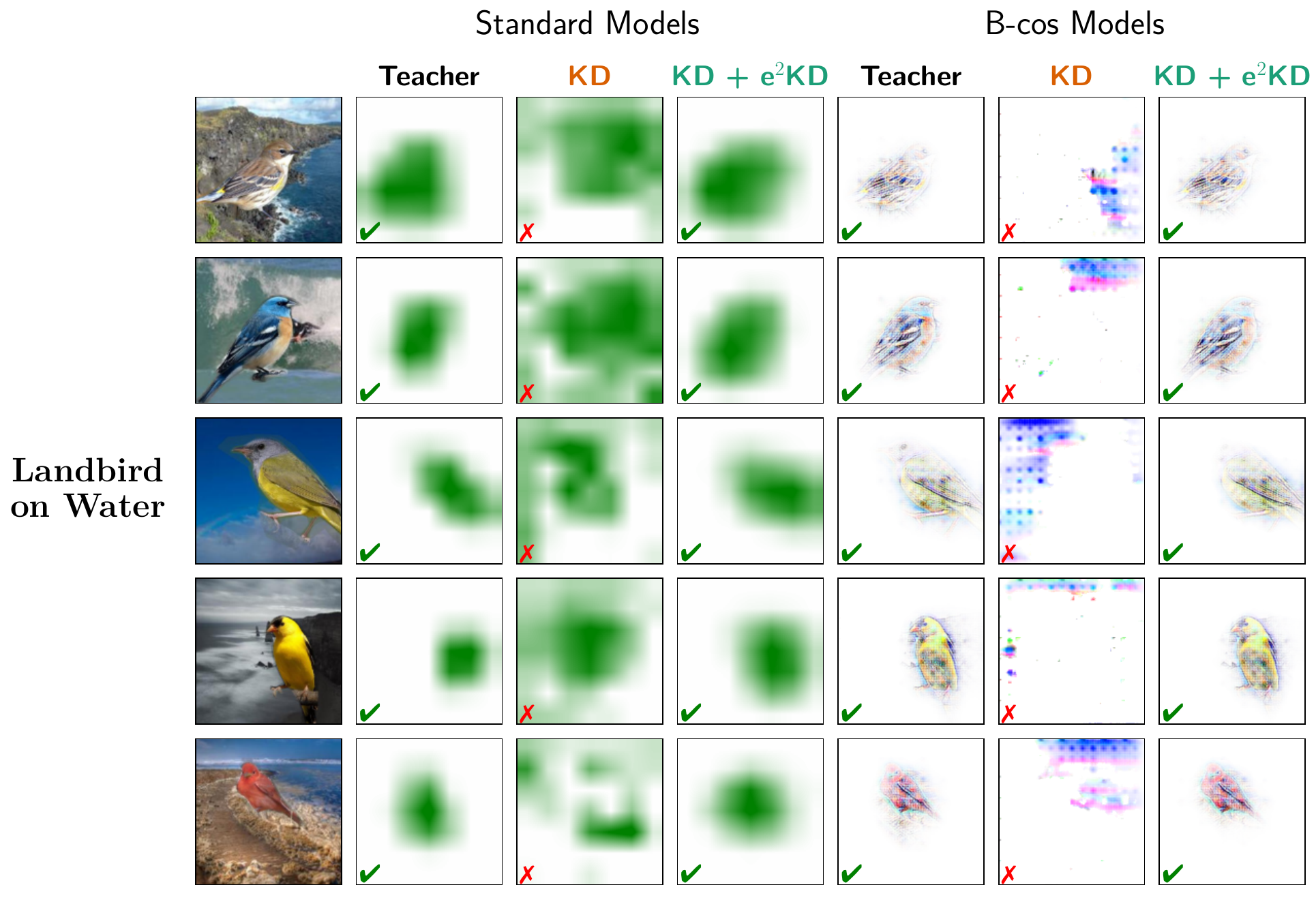}
    \end{subfigure}
    
    \vspace{1em}
    \begin{subfigure}[c]{\textwidth}
    \centering
    \includegraphics[width=0.99\textwidth]{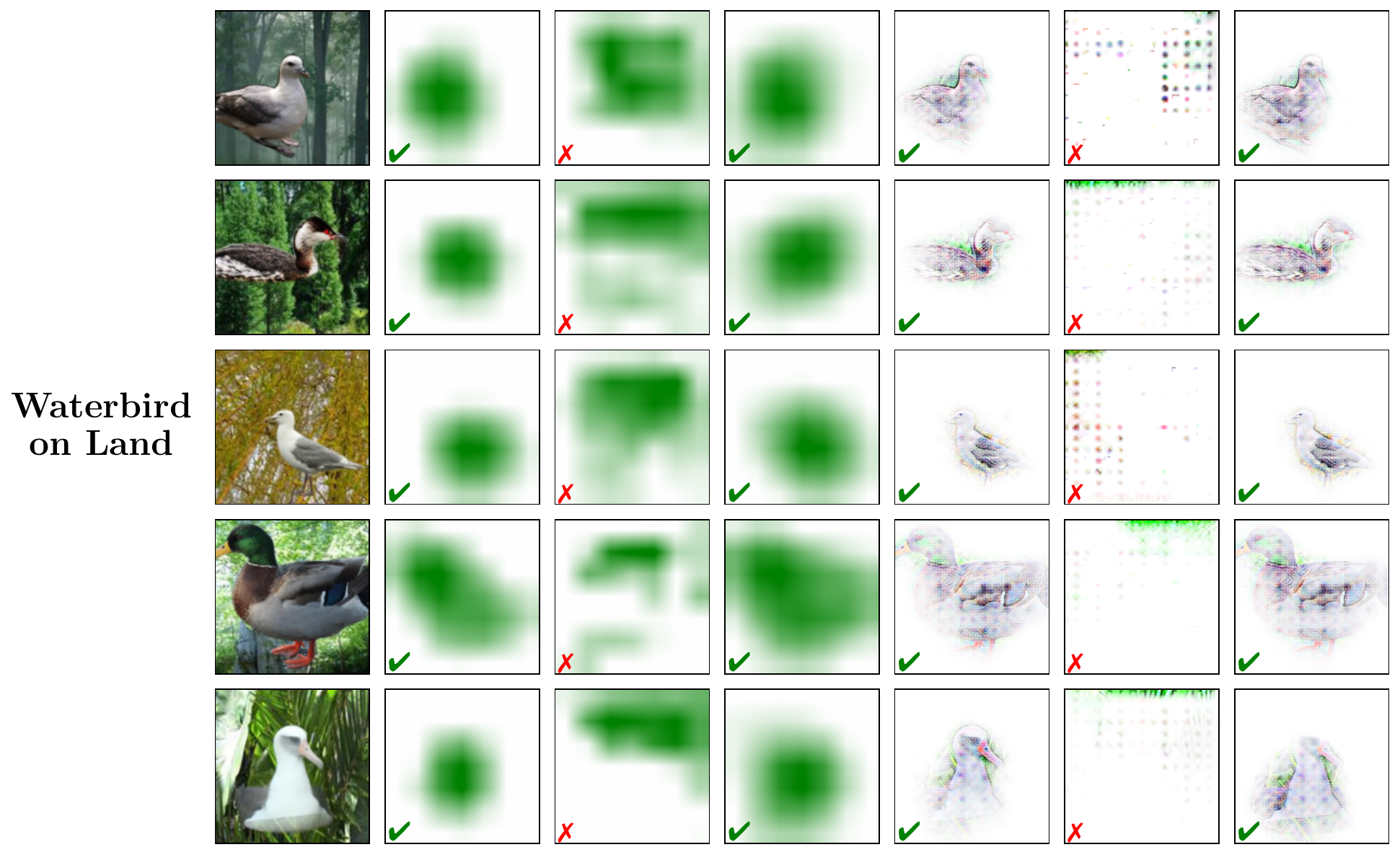}
    \end{subfigure}
    \vspace{-1em}
    \caption{\textbf{Out-of-distribution samples for distillation on biased data using the \waterbirds dataset}. We show explanations for  standard (cols.~2-4) and \bcos models (cols.~5-7), for the out-of-distribution groups `Landbird on Water' (\textbf{top half}) and `Waterbird on Land' (\textbf{bottom half}). \eKD (col.~4 and 7) is effective in preserving the teacher's focus (col.~2 and 5), %
    leading to higher robustness to distribution shifts than when training students via vanilla KD. Correct and incorrect predictions marked by {\color{forestgreen}\ding{51}} and {\color{red}\ding{55}} respectively.}
    \label{fig:wb:qual_ood_supp}
\end{figure}

\clearpage

\subsection{Maintaining Focused Explanations}
\label{supp:sec:qualitative:voc}

In this section we provide additional qualitative examples for experiments on \voc (see \cref{sec:results:voc}). We provide samples for all of the 20 classes in the \voc dataset (every row in \cref{fig:voc:bcos_supp_first,fig:voc:bcos_supp_second}). Across all classes we ovserve that the student trained with \eKD maintains focused explanations on the class-specific input-features, whereas the student trained with vanilla KD may often focus on the background.
\begin{figure}[h!]
    \centering
    \vspace{-15pt}
    \includegraphics[width=0.98\textwidth]{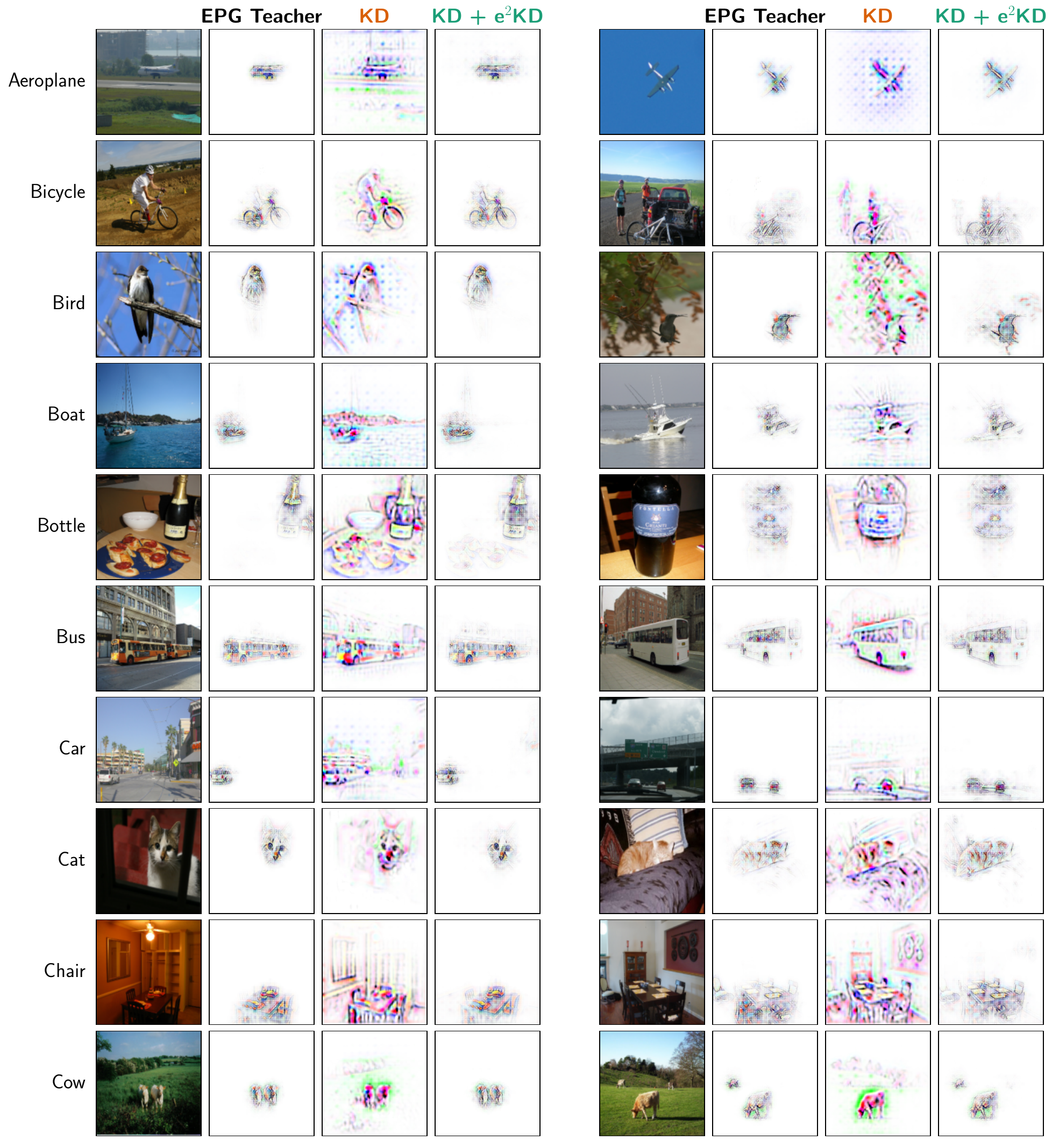}
    \vspace{-5pt}
    \caption{\textbf{Maintaining focused explanations (classes 1-10)}: Similar to \cref{fig:bcos-voc} in the main paper, here we show qualitative difference of explanations. Each row shows two samples per class (for classes 11-20 see \cref{fig:voc:bcos_supp_second}). We find that explanations of the student trained with \eKD (subcol.~4 on both sides) is significantly closer to the teacher's (subcol.~2), whereas vanilla KD students also focus on the background (subcol.~3). Samples were drawn from the test set with all models having correct predictions.}
    \label{fig:voc:bcos_supp_first}
\end{figure}

\begin{figure}[h!]
    \centering
    \includegraphics[width=0.99\textwidth]{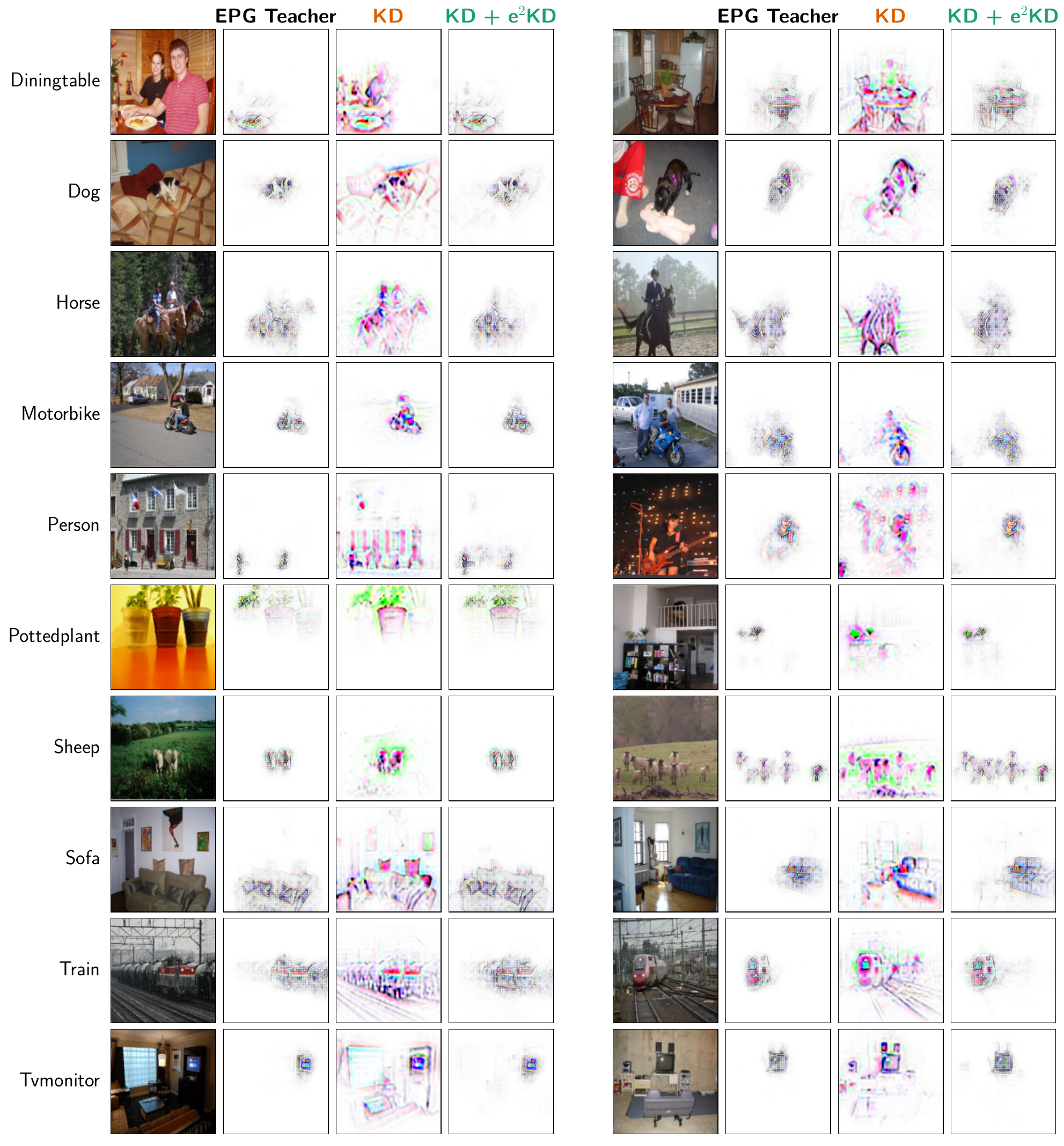}
    \caption{\textbf{Maintaining focused explanations (classes 11-20)}: Similar to \cref{fig:bcos-voc} in the main paper, here we show qualitative difference of explanations. Each row shows two samples per class (for classes 1-10 see \cref{fig:voc:bcos_supp_first}). We find that explanations of the student trained with \eKD (subcol.~4 on both sides) is significantly closer to the teacher's (subcol.~2), whereas vanilla KD students also focus on the background (subcol.~3). Samples were drawn from test set with all models having correct predictions.}
    \label{fig:voc:bcos_supp_second}
\end{figure}
\clearpage

\subsection{Distilling Architectural Priors}
\label{supp:sec:qualitative:vit}
\vspace{-8pt}
We provide additional qualitative samples for \cref{sec:result:priors} in the main paper, where we distill a \bcos CNN to a \bcos ViT. Looking at \cref{fig:vit:bcos_supp}, one can immediately observe the difference in interpretability of the \bcos ViT explanations when trained with \eKD vs.~vanilla KD. Following the discussion in \cref{fig:vit_results}, one can see that the ViT trained with \eKD, similar to its CNN Teacher, gives consistent explanations under shift, despite its inherent tokenization, whereas the explanations from vanilla KD significantly differ (compare odd rows to the one below).
\begin{figure}[!h]
    \vspace{-0.5cm}
    \centering
    \includegraphics[width=0.98\textwidth]{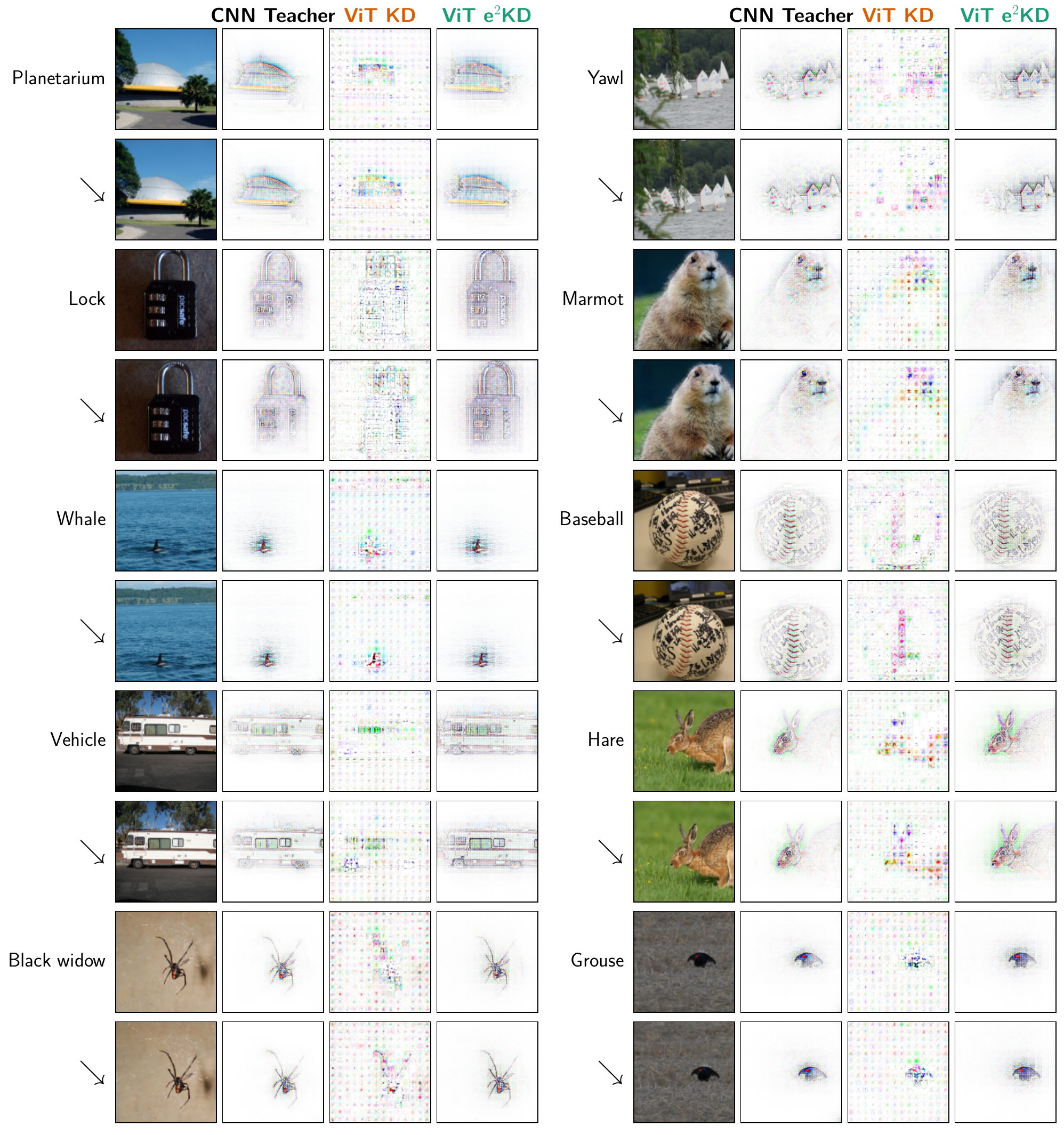}
    \caption{\textbf{Qualitative results on distilling \bcos \densenet-169 to \bcos ViT$_\text{Tiny}$}. The explanations of the \eKD ViT student (subcols.~4) are significantly more interpretable than vanilla KD student (subcols.~3), and very close to the teacher's explanations (subcols.~2). We also shift images diagonally to the bottom right by 8 pixels and show the explanations for the same class (rows indicated by $\searrow$). 
    The explanations of the ViT student trained with \eKD are shift-equivariant, whereas the ViT student from vanilla KD is sensitive to such shifts and the explanations change significantly.
    }
    \label{fig:vit:bcos_supp}
    \vspace{-0.8cm}
\end{figure}

\clearpage
\section{Additional Quantitative Results}
\label{supp:sec:quantitative}

\subsection{Reproducing previously reported results for prior work.}
\label{supp:sec:quantitative:oldrecipe}
Since we use prior work on new settings, namely \imagenet with limited data, Waterbirds-100, and \bcos models, we %
reproduce the performance reported in the original works in the following. %
In particular, in \cref{tbl:imnt:normal:reproduce}, we %
report the results obtained by the training `recipes' from by prior work to validate our implementation and enable better comparability with previously reported results.

For this section, we distilled the standard \resnet-34 teacher to a \resnet-18 on \imagenet for 100 epochs, with an initial learning rate of 0.1, decayed by 10\% every 30 epochs. We used SGD with momentum of 0.9 and a weight-decay factor of 1e-4. For AT, ReviewKD, and CRD, we followed the respective original works and employed weighting coefficients of $\lambda \myin \{1000.0, 1.0, 0.8\}$ respectively. For \catkd, we used $\lambda \myin \{1.0, 5.0, 10.0\}$ after identifying this as a reasonable range in preliminary experiments. Here we also, similar to \cref{supp:sec:impl:training}, used a held-out validation set from the training samples.

Following \citeApp{kd}, we also report the results for KD in which the cross-entropy loss with respect to the ground truth labels is used in the loss function. We were able to reproduce the reported numbers by a close margin. Our numbers are also comparable to the torchdistill's reproduced numbers \citeS{torchdistillS}, see \cref{tbl:imnt:normal:reproduce}. We see that \eKD again improves both accuracy and agreement of vanilla KD (agreement 80.2$\rightarrow$80.5). Similar gains are also observed for CRD (agreement 78.4$\rightarrow$79.2). Also note that the vanilla KD baseline significantly improves once we used the longer training recipe from \citeApp{consistency} in \cref{tbl:imnet:normal} in the main paper (accuracy 70.6$\rightarrow$71.8; agreement 80.2$\rightarrow$81.2). 

\begin{table}[h]
\caption{\textbf{Distilling Standard \resnet-34 to \resnet-18 for reproducing prior work.} We verify our implementation of prior work by distilling them in the 100 epoch setting used in \protect\citeS{attentionS,reviewkdS,catkdS,tian2019crdS}. We observe that our accuracy is very close to the reported one and the reproduced numbers by torchdistill \protect\citeS{torchdistillS}. We also see that \eKD, similar to \cref{tbl:imnet:normal}, improves accuracy and agreement of vanilla KD and CRD.}
\centering
\begin{tabular}{l@{\hskip15pt}c@{\hskip20pt}c@{\hskip20pt}c@{\hskip20pt}c}
% \toprule
\shortstack[c]{\bf Standard Models\\ Teacher ResNet-34 \\{Accuracy} 73.3\%} &  \shortstack[c]{Accuracy}   &  \shortstack[c]{Agreement} &  \shortstack[c]{Reported\\Accuracy} & \shortstack[c]{torchdistill\phantom{p}\\Accuracy}    \\
\midrule
KD \citeApp{kd} {\scriptsize{(with cross-entropy)}}  &           71.0 &              79.7 &      70.7         &         71.4           \\
AT \citeApp{attention}                               &           70.2 &              78.3 &      70.7         &           70.9          \\
ReviewKD \citeApp{reviewkd}                          &           71.6 &              80.1 &      71.6         &           71.6          \\
\catkd \citeApp{catkd}                               &           71.0 &              80.1 &      71.3         &             -            \\
\midrule
KD                           &           70.6 &              80.2 &       -            &             -            \\
\bf+ \eKD \scriptsize (GradCAM)   &   \bf70.7 &      \bf80.5 &       -           &             -            \\
   &    \bf{\scriptsize{\color{forestgreen}\makebox[.75em][c]{+}\makebox[1.75em][r]{ 0.1 }}}&    \bf{\scriptsize{\color{forestgreen}\makebox[.75em][c]{+}\makebox[1.75em][r]{ 0.3 }}}&   &  \\
\midrule
CRD \citeApp{tian2019crd}            &           70.6 &              78.4 &      71.2         &         70.9           \\ 
\bf+ \eKD \scriptsize (GradCAM)   &       \bf 71.0 &          \bf 79.2 &       -           &           -            \\
   &    \bf{\scriptsize{\color{forestgreen}\makebox[.75em][c]{+}\makebox[1.75em][r]{ 0.4 }}}&    \bf{\scriptsize{\color{forestgreen}\makebox[.75em][c]{+}\makebox[1.75em][r]{ 0.8 }}}&   &  \\
\end{tabular}
\label{tbl:imnt:normal:reproduce}
\vspace{-25pt}
\end{table}

\clearpage
\subsection{Full results on \wbirdsshort\ --- \bcos models}
In this section we provide complete quantitative results on \waterbirds \citeS{wb100S,galsS} for \bcos models. In \cref{fig:wb:bcos_supp}, we report in-distribution and out-of-distribution accuracy and agreement. Similar to what was observed for standard models in \cref{fig:wb:ood} of the main paper, here we again observe that all models are performing well on in-distribution data (lowest test-accuracy is 94.9\% for \bcos models). Nevertheless, \eKD (with \bcos explanations) is again consistently providing gains over vanilla KD. More importantly however, for the out-of-distribution samples, we observe that \eKD offers even larger accuracy and agreement gains over vanilla KD for both \bcos (\cref{fig:wb:bcos_supp}) and standard (\cref{fig:wb:ood}, main paper) models. 
Corresponding qualitative results, for the 700 epoch experiments, can be found in \cref{fig:wb:qual_id_supp}, for in-distribution, and \cref{fig:wb:qual_ood_supp} for out-of-distribution samples.
\label{supp:sec:quantitative:waterbirds}
\begin{figure*}[h]
    \centering
    \begin{subfigure}[c]{\textwidth}
    \centering
    \includegraphics[width=\textwidth]{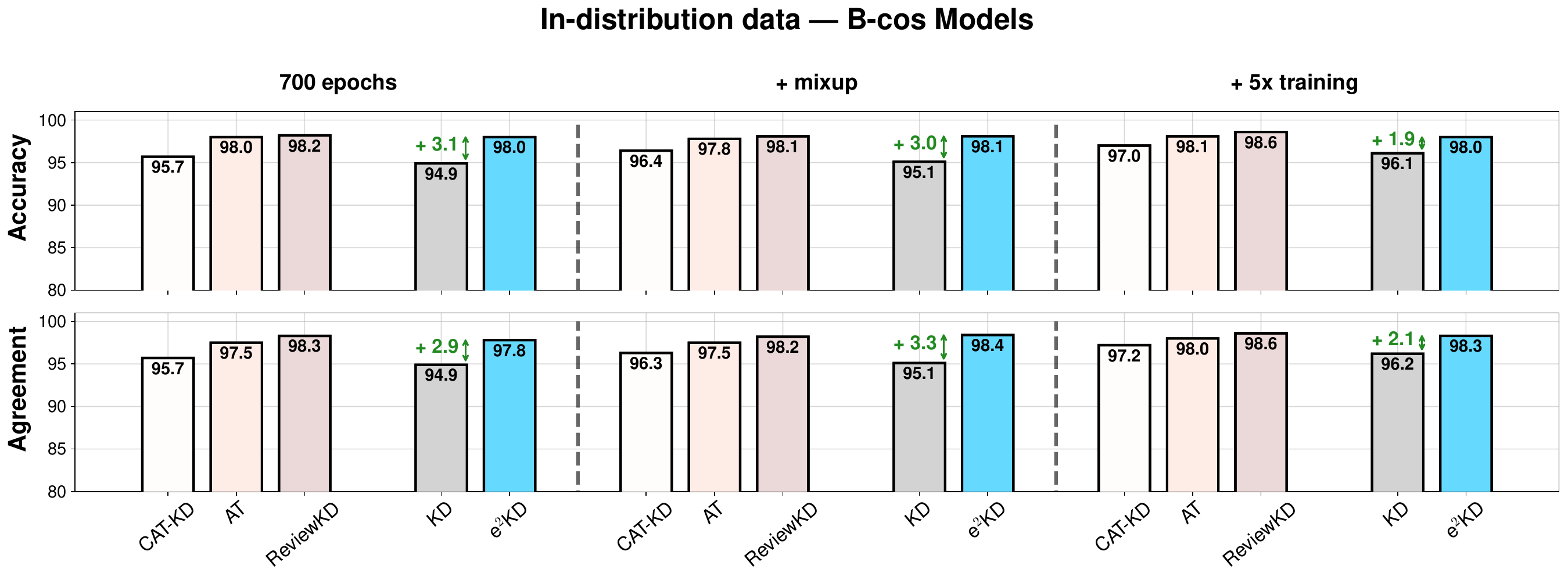}
    \caption{In-distribution --- \bcos Teacher Acc.~: 98.8\%}
    \end{subfigure}
    
    \vspace{2em}
    
    \begin{subfigure}[c]{\textwidth}
    \centering
    \includegraphics[width=\textwidth]{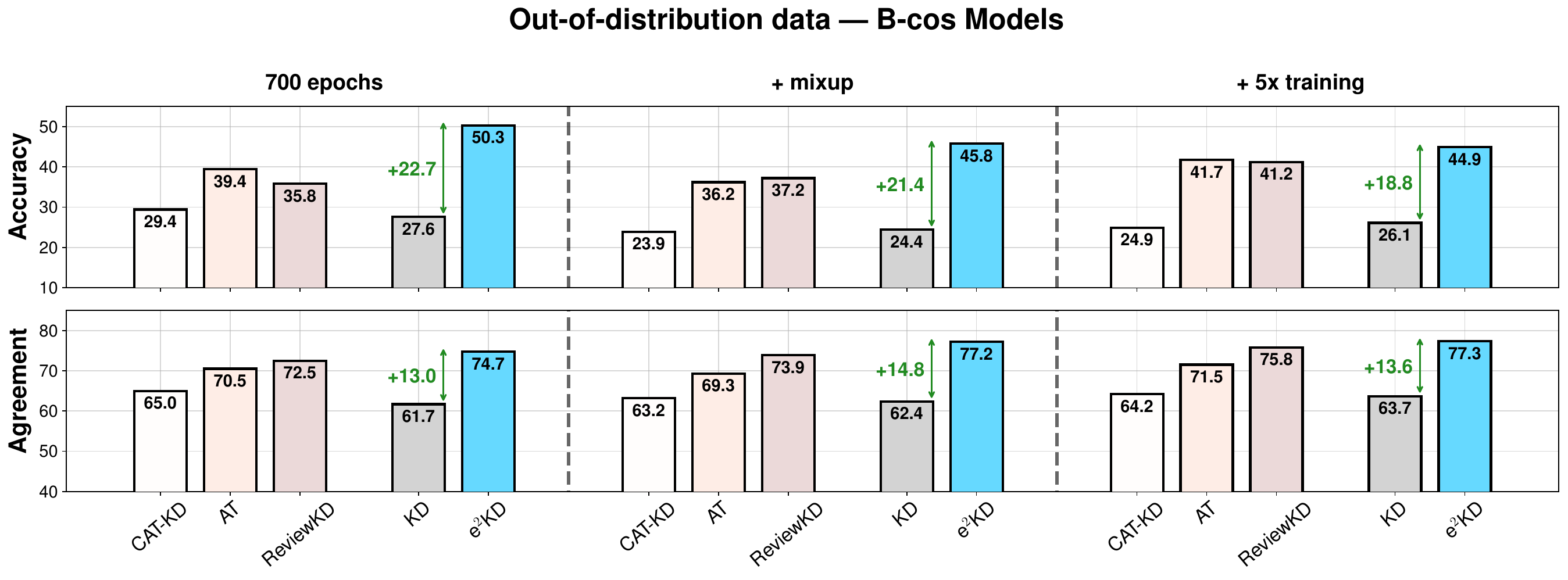}
    \caption{Out-of-distribution --- \bcos Teacher Acc.~: 55.2\%}
    \end{subfigure}
    \caption{
    \textbf{Results for \textit{\bcos} models on \waterbirds.} We show accuracy and agreement on both \emph{in-distribution} \textbf{(top)} and \emph{out-of-distribution} \textbf{(bottom)} test samples when distilling from \bcos ResNet-50 teachers to \bcos ResNet-18 students with various KD approaches. As for out-of-distribution data, we find significant and consistent gains in accuracy and agreement (similar to \cref{fig:wb:ood} for standard models). 
    }
    \vspace{-20pt}
    \label{fig:wb:bcos_supp}
\end{figure*}

\clearpage

\subsection{Detailed Comparison with respect to \catkd}
As mentioned in \cref{sec:related-work} in the main paper, works such as \catkd~\citeApp{catkd} have explored using an explanation method (\eg Class Activation Maps (CAM)~\citeApp{cam}) in the context of knowledge distillation, though without having faithfulness of the distillation as the primary focus. This has resulted in design choices such as down-sampling the explanations to $2\times2$. On the contrary, \eKD is designed towards promoting faithful distillation, by simply matching the explanations. While in this work we explored the benefits of \eKD under both \gradcam~\citeApp{gradcam} explanations for standard models and \bcos~\citeApp{bcos} explanations for \bcos models~\citeS{bcosS,bcosv2S}, since \gradcam and CAM are very similar explanation methods, yet in \cref{tbl:imnet:normal} \eKD significantly outperforms \catkd, in this section we ablate over all of our differences with respect to \catkd.

Specifically, we evaluate the impact of design choices over the limited-data setting for \imagenet dataset (similar to 50- and 200-Shot columns in \cref{tbl:imnet:normal} in the main paper). In \cref{tbl:supp:catkd-ablate} we ablate over resolution of explanation maps (whether to have down-sampling operation from \citeApp{catkd}), choice of explanation method (\cam\xspace vs. \gradcam), and other differences such as taking explanations for all or only teacher's prediction and using cross-entropy w.r.t. labels instead of \cref{eq:vanillakd}. Here we use a coefficient of 10 for the explanation loss. As per the implementation of \catkd~\citeApp{catkd}, the CAM explanations between teacher student are normalized and regularized to be similar with Mean Squared Error loss. This is equivalent to cosine similarity that we use in \eKD, except for a constant  $\frac{2}{H \times W}$ factor, with $(H, W)$ being the resolution of explanation maps, that is applied to the gradient from the explanation loss in \catkd. Therefore, we also re-scale the loss coefficient after changing the resolution of the  maps (3rd row in \cref{tbl:supp:catkd-ablate}). 

In \cref{tbl:supp:catkd-ablate}, we observe that \emph{not} reducing the resolution of CAMs to $2\times2$, results in consistent (accuracy | agreement) gains in both 50-Shot (32.2$\rightarrow$35.0 | 34.5$\rightarrow$37.3) and 200-Shot (55.7$\rightarrow$57.4 | 60.7$\rightarrow$63.0) cases on \imagenet. As one would expect, both CAM and \gradcam explanations result in similar numbers, since they are essentially the same explanation, up to a ReLU operation in \gradcam. 
\label{supp:sec:quantitative:cat-kd}

\begin{table}[h]
\caption{\text{Ablation with respect to \catkd.} Here we disentangle the differences between \eKD and \catkd, with every row bringing it closer to \eKD's setting. We distill a standard \resnet-34 to standard \resnet-18 on a subset of \imagenet and evaluate on the complete test set.}
\centering
\begin{tabular}{l@{\hskip8pt}c c @{\hskip20pt} c c}
\shortstack[c]{\bf Standard Models\\ Teacher \resnet-34:} Accuracy 73.3\%

&  \multicolumn{2}{c}{\bf 50 Shot\phantom{sssss}}   &  \multicolumn{2}{c}{\bf 200 Shot \phantom{s}}  \\
\midrule
CAT-KD                                           &           32.2       &             34.5 &      55.7         &         60.7    \\
+ Remove $2\times2$ down-sampling (use $7\times7$ maps)                &           24.4       &             25.9 &      46.1         &         49.5    \\
+ Re-scale the explanation loss coefficient ($\times \frac{49}{4}$) &    35.0       &             37.3 &      57.4         &         63.0    \\
+ Replace CE with KL-Div., use Top-1 Explanation &           53.1       &             59.3 &      63.7         &         72.6    \\
+ Replace CAM with GradCAM                       &           52.9       &             59.2 &      64.1         &         73.2   \\
\end{tabular}
\label{tbl:supp:catkd-ablate}
\vspace{-1cm}
\end{table}
\clearpage
\section{Implementation Details}
\label{supp:sec:impl}

In this section, we provide additional implementation details. In \cref{supp:sec:impl:training}, we provide a detailed description of our training setup, including hyperparameters used in each setting. In \cref{supp:sec:impl:bcos-adaptation}, we describe how we adapt prior approaches that were proposed for conventional deep neural networks to \bcos models. Code for all the experiments will be made available.

\subsection{Training Details}
\label{supp:sec:impl:training}

In this section, we first provide the general setup which, unless specified otherwise, is shared across all of our experiments. Afterwards, we describe dataset-specific details in \cref{supp:sec:impl:training:imagenet,supp:sec:impl:training:waterbirds,supp:sec:impl:training:voc}, for each dataset and experiment.

\myparagraph{Standard Networks.}
As mentioned in \cref{sec:results} in the main paper, we follow the recipe from \citeApp{consistency}. For standard models, we use the AdamW optimizer \citeApp{kingma2014adam} with a weight-decay factor of $10^{-4}$ and a cosine-annealing learning-rate scheduler \citeApp{loshchilov2017sgdr} with an initial learning-rate of 0.01, reached with an initial warmup for 5 epochs. We clip gradients by norm at 1.0.

\myparagraph{\bcos Networks.}  We use the latest implementations for \bcos models~\citeApp{bcosv2}. Following \citeS{bcosS, bcosv2S}, we use the Adam optimizer \citeApp{kingma2014adam} and do not apply weight-decay. We use a cosine-annealing learning-rate scheduler  \citeApp{loshchilov2017sgdr} with an initial learning-rate of $10^{-3}$, reached with an initial warmup for 5 epochs. Following \citeApp{bcosv2}, we clip gradients using adaptive gradient clipping (AGC) \citeApp{agc-pmlr-v139-brock21a}. Unless specified otherwise, across all models and datasets we use random crops and random horizontal flips as data augmentation during training, and at test time we resize the images to 256 (along the smaller dimension) and apply center crop of (224, 224). We use PyTorch \citeApp{paszke2019pytorch} and PyTorch Lightning \citeApp{Falcon_PyTorch_Lightning_2019} for all of our implementations.

\subsubsection{ImageNet Experiments.}
\label{supp:sec:impl:training:imagenet}

For experiments on the full \imagenet dataset~\citeApp{imagenet}, we use a batch size of 256 and train for 200 epochs. For limited-data experiments we keep the number of steps same across both settings (roughly 40\% total steps compared to full-data): when using 50 shots per class, we set the batch size to 32 and train for 250 epochs, and when having 200 shots, we use a batch size of 64 and train for 125 epochs. We use the same randomly selected shots for all limited-data experiments. For experiments on unrelated data (\cref{tbl:sun2im} (right) in the main paper), following \citeApp{consistency}, we used equal number of training steps for both SUN$\rightarrow$IMN (125 epochs with a batch size of 128) and IMN$\rightarrow$SUN (21 epochs with a batch size of 256).

For the pre-trained teachers, we use the Torchvision checkpoints\footnote{\href{https://pytorch.org/vision/stable/models.html}{https://pytorch.org/vision/stable/models.html}} \citeApp{torchvision2016} for standard models and available checkpoints for B-cos models\footnote{\href{https://github.com/B-cos/B-cos-v2}{https://github.com/B-cos/B-cos-v2}} \citeApp{bcosv2}. For all \imagenet experiments, we pick the best checkpoint and loss coefficients based on a held-out subset of the standard train set, which has 50 random samples per class. The results are then reported on the entire official validation set. We use the following parameters for each method:\\[.5em]
    
\noindent\begin{minipage}[t]{.5\textwidth}
\phantom{something}
\begin{tabular}{ll}
\multicolumn{2}{l}{\bf Standard Networks (\cref{tbl:imnet:normal})}\\
     KD &$\tau \in [1, 5]$\\
     \phantom{+}\eKD &$\tau \in [1, 5]$,  $\lambda \in [1, 5, 10]$\\
     AT& $\lambda\in [10, 100, 1000, 10000]$\\
     ReviewKD& $\lambda\in [1, 5]$\\
     \catkd &$\lambda\in [1, 5, 10]$\\
     CRD &$\lambda\in [0.8]$\\
     \phantom{+}\eKD &$\lambda_\text{CRD} \in [0.8]$,  $\lambda \in [1, 5, 10]$\\
\end{tabular}
\end{minipage}
\begin{minipage}[t]{.5\textwidth}
\phantom{something}
\begin{tabular}{ll}
\multicolumn{2}{l}{\bf\bcos \resnet-34 Teacher (\cref{tbl:imnet:bcos})}\\
     KD &$\tau \in [1, 5]$\\
     \eKD &$\tau \in [1, 5]$, $\lambda \in [1, 5]$\\
     AT &$\lambda \in [1, 10, 100, 1000]$\\
     ReviewKD &$\lambda \in [1, 5]$\\
     \catkd &$\lambda \in [1, 5]$
\end{tabular}
\end{minipage}\\[.5em]

\noindent\begin{minipage}{.50\textwidth}
% \phantom{something}
% \begin{center}
\begin{tabular}{ll}
\multicolumn{2}{l}{\bf\bcos \densenet-169 Teacher}\\
\bf (\cref{tbl:imnet:bcos:densenet})\\
    KD  &$\tau \in [1, 5]$\\
    \eKD &$\tau \in [1, 5]$, $\lambda \in [1, 5, 10]$\\
    \snowflake KD& $\tau \in [1, 5]$\\
    \snowflake \eKD &$\tau \in [1, 5]$, $\lambda \in [0.2, 1, 5, 10]$\\
\end{tabular}
% \end{center}
\end{minipage}\begin{minipage}{.45\textwidth}
% \begin{center}
\begin{tabular}{ll}
\multicolumn{2}{l}{\bf Unrelated Data (\cref{tbl:sun2im}, Right)}\\
\multicolumn{2}{l}{\bcos \densenet-169 Teacher}\\
    KD& $\tau  \in [1, 5]$\\
    \eKD& $\tau  \in [1, 5]$, $\lambda \in [1, 5, 10]$\\
    \ \\
    \ \\
\end{tabular}
% \end{center}
\end{minipage}
% \hfill

\vspace{0.5cm}
For ViT students we trained for 150 epochs, and following \citeApp{bcosv2}, we used 10k warmup steps, and additionally used RandAugment\citeApp{NEURIPS2020_d85b63ef} with magnitude of 10.
\begin{center}
\noindent\begin{minipage}{\textwidth}
\centering
\begin{tabular}{ll}
\multicolumn{2}{l}{\bf \bcos \densenet-169 to ViT Student (\cref{fig:vit_results})}\\
     KD& $\tau \in [1, 5]$\\
     \eKD& $\tau \in [1]$, $\lambda \in [1, 5, 10]$
\end{tabular}
\end{minipage}
\end{center}

\subsubsection{Waterbirds Experiments.}
\label{supp:sec:impl:training:waterbirds}

For the Waterbirds \citeApp{wb100} experiments, we use the 100\% correlated data generated by \citeApp{gals} (\ie Waterbirds-100). We use the provided train, validation and test splits. Since the data is imbalanced (number of samples per class significantly differ), within each sweep we pick the last-epoch checkpoint with best \emph{overall} validation accuracy (including both in-distribution and out-of-distribution samples). We use batch size of 64. For experiments with MixUp, we use $\alpha$=1. For applying AT and ReviewKD between the \resnet-50 teacher and \resnet-18 student, we used the same configuration from a \resnet-34 teacher, since they have the same number of blocks.

The pre-trained guided teachers were obtained from \citeApp{modelguidance}. The standard \resnet-50 teacher had 99.0\% and 61.2\% in-distribution and out-of-distribution accuracy, and the \bcos \resnet-50 teacher had 98.8\% and 55.1\% respectively.

We tested the following parameters for each method:

\begin{center}
\begin{minipage}{.5\textwidth}
\begin{center}
\begin{tabular}{ll}
\multicolumn{2}{l}{\bf Standard models (\cref{fig:wb:ood})}\\
     KD& $\tau \in [1, 5]$\\ 
     \eKD& $\tau \in [1, 5]$, $\lambda \in [1, 5, 10, 15]$\\
     AT& $\lambda \in [10, 100, 1000]$\\
     ReviewKD& $\lambda \in [1, 5, 10, 15]$\\
     \catkd & $\lambda \in [1, 5, 10, 15]$
\end{tabular}
\end{center}
\end{minipage}
\begin{minipage}{.45\textwidth}
\begin{center}
\begin{tabular}{ll}
\multicolumn{2}{l}{\bf\bcos models (\cref{supp:sec:quantitative:waterbirds} )}\\
     KD& $\tau \in [1, 5]$\\
     \eKD& $\tau \in [1]$, $\lambda \in [1, 5, 10]$\\
     AT& $\lambda \in [10, 100, 1000]$\\
     ReviewKD& $\lambda \in [1, 5, 10]$\\
     \catkd& $\lambda \in [1, 5, 10]$
\end{tabular}
\end{center}
\end{minipage}
\end{center}

For the results in \cref{tab:wb_variety} in the main paper, we trained ConvNeXt$_\text{Tiny}$~\citeApp{liu2022convnet}, EfficientNetV2$_\text{Small}$~\citeApp{pmlr-v97-tan19a}, MobileNetV2~\citeApp{Sandler_2018_CVPR}, and ShuffleNetV2$_{\times{0.5}}$~\citeApp{Ma_2018_ECCV} under the 700-Epoch recipe.

\subsubsection{PASCAL VOC Experiments.}
\label{supp:sec:impl:training:voc}

We use the 2012 release of PASCAL VOC dataset \citeApp{pascal-voc-2012}. We randomly select 10\% of the train samples as validation set and report results on the official test set. We use batch size of 64 and train for 150 epochs. The pre-trained guided teachers were obtained from \citeApp{modelguidance}. Since we are using VOC as a \emph{multi-label} classification setting, we replace the logit loss from \cref{eq:vanillakd} in the main paper with the logit loss recently introduced by \citeApp{yang2023multi}: 
\begin{align}
\begin{split}
\label{eq:mld_supp}
         \mathcal L_\mathit{MLD} &= \tau \sum_{j=1}^{c} \KL\left( \left[\psi_{j}\left(\dfrac{z_{T}}{\tau}\right), 1 - \psi_{j}\left(\dfrac{z_{T}}{\tau}\right)\right] || \left[\psi_{j}\left(\dfrac{z_{S}}{\tau}\right), 1 - \psi_{j}\left(\dfrac{z_{S}}{\tau}\right)\right]\right)\;.
\end{split}
\end{align}
Here, $\psi$ is the sigmoid function and $\left[.,.\right]$ concatenates values into a vector. Note that the original loss from \citeApp{yang2023multi} does not have a temperature parameter $\tau$ (\ie $\tau = 1$). For consistency with other experiments, here we also included a temperature factor.
When reporting the final results on the test set, we resize images to (224, 224) and do not apply center crop.
For the EPG and IoU metrics, we use the implementation from \citeApp{modelguidance}. For the IoU metric, we use threshold of 0.05. We tested the following parameters for each method:

\begin{center}
\begin{tabular}{ll}
\multicolumn{2}{l}{\bf \bcos models (\cref{tbl:voc}, Left)}\\
     KD & $\tau \in [1, 5]$,\\
     \eKD & $\tau \in [1, 5]$, $\lambda \in [1, 5, 10]$     
\end{tabular}
\end{center}

\subsection{Adapting Prior Feature-based Methods for B-cos Models}
\label{supp:sec:impl:bcos-adaptation}

While prior feature-based KD methods have been mainly introduced for conventional networks, in \cref{tbl:imnet:bcos} we additionally tested them on \bcos networks. We applied them with the same configuration that they were originally introduced as for \resnet-34 \citeApp{he2016deep} teacher and \resnet-18 student, with minor adjustments. Specifically, since \bcos networks also operate on negative subspace, we did not apply ReLU on the intermediate tensors in AT. For ReviewKD, since the additional convlution and norm layers between the teacher and student are only needed to convert intermediate representations, we used standard convolution and BatchNorm and not B-cos specific layers. For AT, ReviewKD, and \catkd we replaced the cross-entropy loss, with the modified binary cross entropy from \citeApp{bcosv2}.

{
\bibliographystyleS{splncs04}
\bibliographyS{supp_bib}
}

\end{document}